\documentclass[journal]{IEEEtran}
\usepackage{verbatim}
\usepackage{amsfonts}

\usepackage{graphicx}
\usepackage{amsmath,amssymb} 
\usepackage{color}
\usepackage{epstopdf}

\usepackage{float}
\usepackage{multirow}
\usepackage{makecell}
\usepackage{color}
\usepackage{graphicx}
\ifCLASSINFOpdf
\else
\fi

\begin{document}
\title{Perceptual underwater image enhancement with deep learning and physical priors}
\author{
Long~Chen,~
Zheheng Jiang,
Lei Tong,
Zhihua Liu,
Aite Zhao,
Qianni Zhang,
Junyu Dong,~
and~Huiyu Zhou~
\thanks{L. Chen, Z. Jiang, L. Tong, Z. Liu and H. Zhou are with School of Informatics, University of Leicester, United Kingdom, e-mail: (lc408,zj53,lt228,zl208,hz143@leicester.ac.uk). H. Zhou is the corresponding author. }
\thanks{A. Zhao and J. Dong are with Department of information science and engineering, Ocean University of China, China, e-mail: (zhaoaite@stu.ouc.edu.cn;dongjunyu@ouc.edu.cn).}
\thanks{Q. Zhang with School of Electronic Engineering and Computer Science, Queen Mary University of London, United Kingdom, e-mail: (qianni.zhang@eecs.qmul.ac.uk).}
\thanks{Manuscript received June 1, 2020; revised xxxxx.}}

\markboth{}%
{Shell \MakeLowercase{\textit{et al.}}: Bare Demo of IEEEtran.cls for IEEE Journals}

\maketitle
\begin{abstract}
Underwater image enhancement, as a pre-processing step to improve the accuracy of the following object detection task, has drawn considerable attention in the field of underwater navigation and ocean exploration. However, most of the existing underwater image enhancement strategies tend to consider enhancement and detection as two independent modules with no interaction, and the practice of separate optimization does not always help the underwater object detection task. In this paper, we propose two perceptual enhancement models, each of which uses a deep enhancement model with a detection perceptor. The detection perceptor provides coherent information in the form of gradients to the enhancement model, guiding the enhancement model to generate patch level visually pleasing images or detection favourable images. In addition, due to the lack of training data, a hybrid underwater image synthesis model, which fuses physical priors and data-driven cues, is proposed to synthesize training data and generalise our enhancement model for real-world underwater images. Experimental results show the superiority of our proposed method over several state-of-the-art methods on both real-world and synthetic underwater datasets.
\end{abstract}
%
\IEEEpeerreviewmaketitle
\section{Introduction}
Underwater object detection (UOD) is of great importance for underwater applications such as ocean exploring and monitoring and autonomous underwater vehicles \cite{b1}. However, underwater images acquired in complicated underwater environments suffer from serious distortion which dramatically degrades the image visibility and affects the detection accuracy of UOD tasks.

In recent years, underwater image enhancement (UIE) technologies \cite{b2, b3}, especially deep learning-based approaches, work as a pre-precessing operation to boost the detection accuracy of UOD tasks by improving the visual quality of underwater images. However, most of the existing strategies consider UIE and UOD tasks as two separate pipelines, whereas the UIE task is evaluated on the visual quality of images while the UOD task is evaluated on the detection accuracy. Separate optimisation of the two tasks results in inconsistency in the rendering of image quality and detection accuracy: These two tasks have different optimisation objectives, leading to different optimal solutions. Moreover, current top-performing deep learning-based UIE methods \cite{b4, b5} are normally trained on synthetic images due to the lack of large and useful training data (i.e., pairs of degraded underwater images and high-quality counterparts). The enhancement models trained on synthetic images cannot always be generalised to underwater scenes because the quality of synthetic images cannot be guaranteed by the existing underwater image synthesis (UIS) methods.

\begin{figure}[htb]
\centering
\includegraphics[height=7cm, width=9cm]{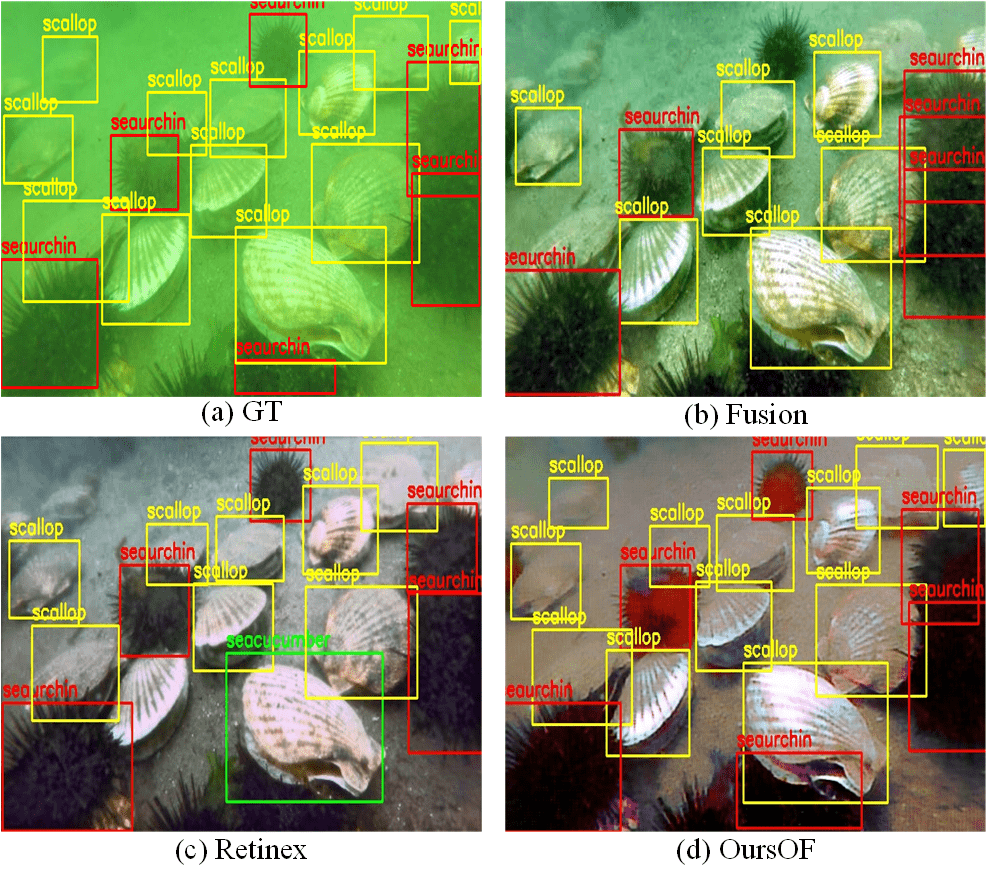}
\caption{Object detection results of the Single Shot MultiBox Detector \cite{b6} after we have applied different UIE algorithms, including (b) Fusion \cite{b7}, (c) Retinex \cite{b8}, and (d) Our proposed object-focused perceptual enhancement model. (a) Raw underwater image with ground-truth annotations.}
\label{fig:overview}
\end{figure}
To address these two concerns, we firstly propose a hybrid underwater image synthesis model to synthesize realistic training data using data-driven cues and physical priors. This enables our enhancement model to generalize well on real-world underwater scenes. Secondly, we propose two detection-perceptual enhancement models, each of which consists of an enhancement model and a detection perceptor. The detection perceptor is well-trained on high-quality in-air images and encodes fine object details and potential detection favouring information of high-quality in-air images. Two perceptual losses are designed to transfer the knowledge encoded in the detection perceptors to the enhancement model in the form of gradients (inference of updating directions). One of the detection perceptors is named patch detection perceptor with a patch perceptual loss and guides the enhancement model to generate patch level visually pleasing images. The other one is named object-focused detection perceptor with an object-focused perceptual loss and guides the enhancement model to generate detection-favouring images. Fig~\ref{fig:overview} shows the object detection results of the same Single Shot MultiBox Detector (SSD) \cite{b6} trained on the enhanced results of different UIE algorithms. The deep detectors trained on the enhanced results of two representative UIE algorithms, i.e., Fusion \cite{b7} and Retinex \cite{b8}, often miss detecting ``noisy" objects or predict incorrect object categories, while our object-focused perceptual enhancement model (denotes as OursOF) can largely improve the detection accuracy of the standard deep detector. The proposed synthesis and the perceptual enhancement models are integrated into a unified framework named HybridDetectionGAN that can work in an end-to-end style. Our contributions can be summarised as follows:
\begin{itemize}
\item We propose two detection perceptual enhancement models, each of which consists of an enhancement model and a detection perceptor. The enhancement model can generate visually pleasing images on the patch-level, while the other one can generate detection-favouring images that help improving the detection accuracy. To our knowledge, this is the first practice on underwater image enhancement, aiming to generate detection-favouring rather than visually pleasing images.
\item We propose a hybrid underwater image synthesis model, which incorporates both physical priors and data-driven cues. The hybrid synthesis model fully takes into account image characteristics such as color distortion, haze-effects and diversity, enabling our perceptual enhancement models to be generalised to handle real-world underwater scenes.
\item The proposed hybrid synthesis and perceptual enhancement models are incorporated into a unified framework named HybridDetectionGAN, and can be jointly optimised in an end-to-end pattern.
\item We conduct extensive evaluations and our proposed perceptual enhancement models outperform several state-of-the art UIE algorithms using both image quality evaluation metrics and task-related detection accuracy metric on both synthetic and real-world underwater datasets.
\end{itemize}

The rest of the paper is organised as follows: Section \ref{sec:relatework} summarises the related works. Section \ref{sec:proposedmethod} describes the proposed hybrid synthesis and perceptual enhancement models. Section \ref{sec:exmperiments} describes the experimental set-up and Section \ref{sec:exmperimentalresults} reports and discusses the experimental results.
\section{Related Work}
\label{sec:relatework}
With the success of deep learning in numerous computer vision tasks, researchers have moved their focus from traditional prior-based methods to data-driven methods in the field of UIS and UIE. In this section, we first summarise previous UIS and UIE methods. Then, we discuss the perceptual loss related to our proposed detection perceptual enhancement model.
\subsection{Underwater Image Synthesis Method}
Due to the lack of training image pairs, some UIS methods are proposed to synthesize distorted underwater images from high quality in-air RGB or RGB-D images, and construct underwater-in-air image pairs as training data. It can be broadly classified into two categories: physical model-based and deep learning-based UIS methods.
\subsubsection{Physical model-based synthesis methods}
Physical model-based UIS methods usually use an underwater image formation model to synthesize underwater images. Several works \cite{b9, b10, b11} followed the underwater image formation model proposed in \cite{b12} and synthesized underwater image datasets which cover 10 Jerlov water types \cite{b13, b14}. The physical model assumes that an underwater image is formed with three optical processes: light absorption, light back-scattering and light forward-scattering. However, the commonly used underwater image formation model \cite{b9, b12} neglects the forward-scattering for simplify and is formulated as:
\begin{equation}
	I_{w} = I_{ab} + I_{bs}
\label{eq:oriphysical}
\end{equation}
where $I_{w}$, $I_{ab}$ and $I_{bs}$ are the underwater image, the image with light absorption and the image with light back-scattering, respectively. The light absorption component $I_{ab}$ is the component of light reflected directly by the target object into the camera, which can be modelled as:
\begin{equation}
	I_{ab} = I_{a}T = I_{a}e^{-\eta^{\lambda}d}, \lambda \in\{r,g,b\}
\end{equation}
where $I_{a}$ is the undistorted in-air image before propagating through the water, $d$ is the distance from the camera to the captured scene and its value can be known from the depth image. $\eta$ is the wavelength-dependent attenuation coefficient. $\lambda$ denotes different channels of an image, including red, green, and blue channels. $T$ denotes the transmission map, indicating how many lights are kept after the absorption process.

A photon of light travels through the water, subject to light scattering which principally builds the characteristics of haze-effects. Denote $I_{bs}$ as the image suffering from light back-scattering, formulated as follows:
\begin{equation}
	I_{bs} = B^{\lambda}(1-T) = B^{\lambda}(1-e^{-\eta^{\lambda}d}), \lambda \in\{r,g,b\}
\end{equation}
where $B^{\lambda}$ denotes the global background light which is a scalar parameter dependent on wavelength.\\

\subsubsection{Deep learning-based synthesis methods}
Recently, Generative Adversarial Networks (GANs) \cite{b15, b16, b4, b5} have been investigated in the underwater image synthesis field due to its successes in image-to-image translation tasks. Li et al \cite{b16} treated the UIS task as an image-to-image translation task and  exploited a single GAN to synthesize underwater images from in-air RGB-D images. Their generator model can be broken down into three stages: (1) light absorption, which simulates the light absorption process by referring to optical priors; (2) light scattering, which simulates the light scattering process using a shallow convolutional neural network without referring to the optical priors; (3) Vignetting, which produces a shading effect on the image corners, caused by certain camera lenses. However, their weakly-supervised synthesis models had to be trained on unpaired images, making it difficult to simulate the detailed image contents such as colors and textures. To alleviate the needs for training image pairs, Fabbri et al. \cite{b5} applied a two-way Cycle-Consistent Adversarial Networks (CycleGAN) \cite{b17}, which allows learning the mutual translation between in-air and underwater domains from unpaired images. The CycleGAN includes two generators, one generator translates in-air images to underwater ones and can be regarded as the image synthesis model, whilst the other one translates underwater images into in-air images which can be regarded as the image enhancement model.

However, both physical model-based and GAN-based UIS methods cannot accurately model the degradation progress of underwater imaging, and result in unsatisfactory synthetic images \cite{b1, b5, b18}. The commonly used physical underwater image formation model can only synthesize 10 Jerlov water types and considers only two factors in the degradation progress, leading to significant errors in the generated images. Moreover, GAN-based methods \cite{b19, b20} encounter the model collapse problem that generates images with monotonous colors and frequent artifacts. Their capability to modelling haze-effects is also limited. Different from previous works, we first improve the physical image formation model, and leverage both physical priors and data-driven cues to a unified hybrid synthesis model to create more realistic underwater images.

\subsection{Underwater Image Enhancement Method}
Underwater image enhancement is an indispensable step to improve the visual quality of underwater images and can be categorised into the following three groups: model-free, physical model based, and deep-learning based methods.
\subsubsection{Model-free enhancement methods}
Model-free UIE methods \cite{b7, b21, b8} aim to adjust image pixel values to improve the visual quality without referring to any physical imaging models. Ancuti et al. \cite{b13} proposed a fusion-based underwater image enhancement method by fusing a contrast enhanced underwater image and a color corrected image in a multi-scale fusion strategy. Fu et al. \cite{b21} presented a two-step approach for underwater image enhancement, which includes a color correction algorithm based on piece-wise linear transformation and a contrast enhancement algorithm. Fu et al. \cite{b8} proposed a Retinex-based method for underwater image enhancement, which consists of color correction, layer decomposition and enhancement. Zhang et al. \cite{b22} extended the Retinex-based method by utilizing the bilateral filter and trilateral filter on the three channels of the image in CIELAB color space. The model-free methods can improve visual quality to some extent, but may accentuate noise, introduce artifacts, and suffer from color distortion.

\subsubsection{Physical model-based enhancement methods}
Physical model-based methods \cite{b23, b24, b25, b26} regard underwater image enhancement as an inverse problem of image degradation. These methods usually establish a physical underwater image degradation model, and then estimate the unknown model parameters from various prior assumptions. Finally, the high quality images can be restored by inverting this degradation process. Li et al. \cite{b23} proposed an underwater dark channel prior (UDCP) by adapting the classical dark channel prior (DCP) \cite{b26} into underwater scenes. Peng et al. \cite{b24} proposed a generalised dark channel prior (GDCP) for image enhancement, which incorporates adaptive color correction into an image formation model. Galdran et al. \cite{b42} also proposed a variant of the dark channel prior algorithm, namely Red Channel, which recovers the lost contrast of an underwater image by restoring the colors associated with short wavelengths. Instead of using the DCP \cite{b26} prior, Li et al. \cite{b27} employed a random forest regression model to estimate the medium transmission of the underwater scenes. Peng et al. \cite{b10} estimate the scene depth via image blurriness and light absorption, and employ the estimated depth to enhance underwater images. Li et al. \cite{b28} proposed an underwater image enhancement method, based on the minimum information loss principle and histogram distribution priors.

\subsubsection{Deep learning-based enhancement methods}
Deep learning-based enhancement methods \cite{b4, b5, b29, b30} usually construct deep neural networks and train them using pairs of degraded underwater images and high-quality counterparts. Li et al. \cite{b2} first synthesized underwater images from the RGB-D in-air image, and then trained a two-stage network for underwater image restoration with the synthetic training data. Liu et al. proposed an underwater color transfer model \cite{b5} based on Cycle-Consistent Adversarial Networks \cite{b17}. Li et al. \cite{b29} employed a gated fusion network architecture to learn three confidence maps used to combine the three input images into an enhanced one.

However, all of the three category enhancement methods only improve the visibility of underwater image while ignoring their influence on the later high-level tasks such as underwater object detection \cite{b31, b32}.
\begin{figure*}[htb]
\centering
\includegraphics[height=6cm, width=18cm]{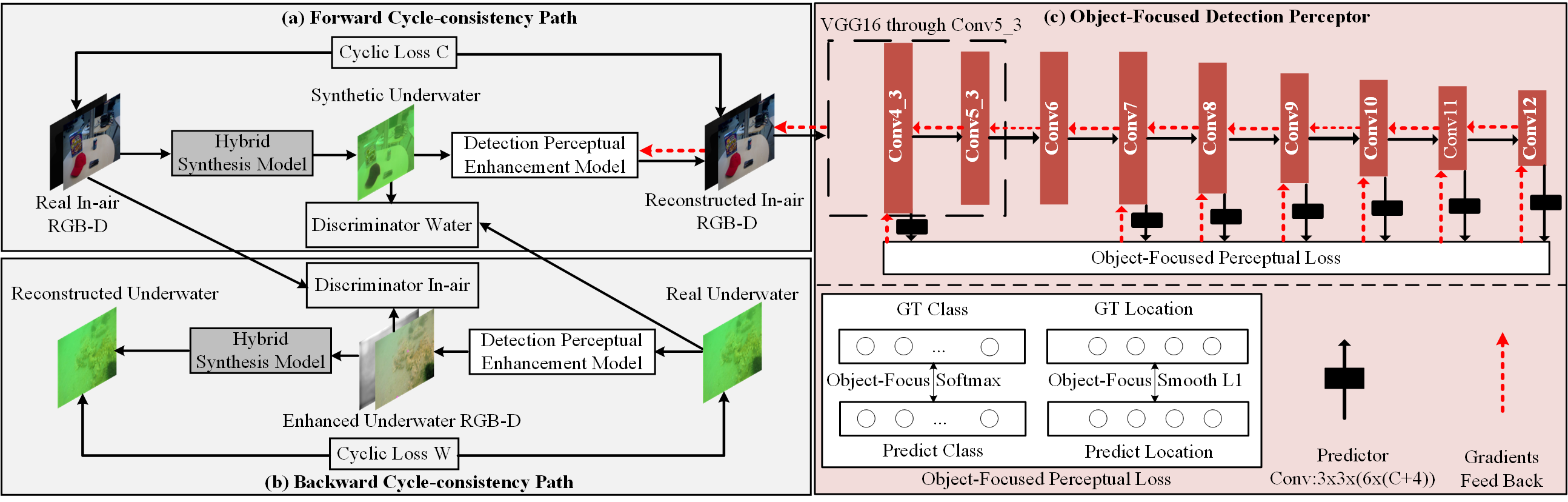}
\caption{The overview of our HybridDetectionGAN. It consists of two cycle-consistency paths (a) and (b) to learn the transformation between underwater and in-air domains from unpaired images. An object-focused detection perceptor (c) guides the enhancement model to generate detection favourable images.}
\label{fig:HybridDetectionGAN}
\end{figure*}
\begin{figure*}[t]
\centering
\includegraphics[height=6.5cm, width=18cm]{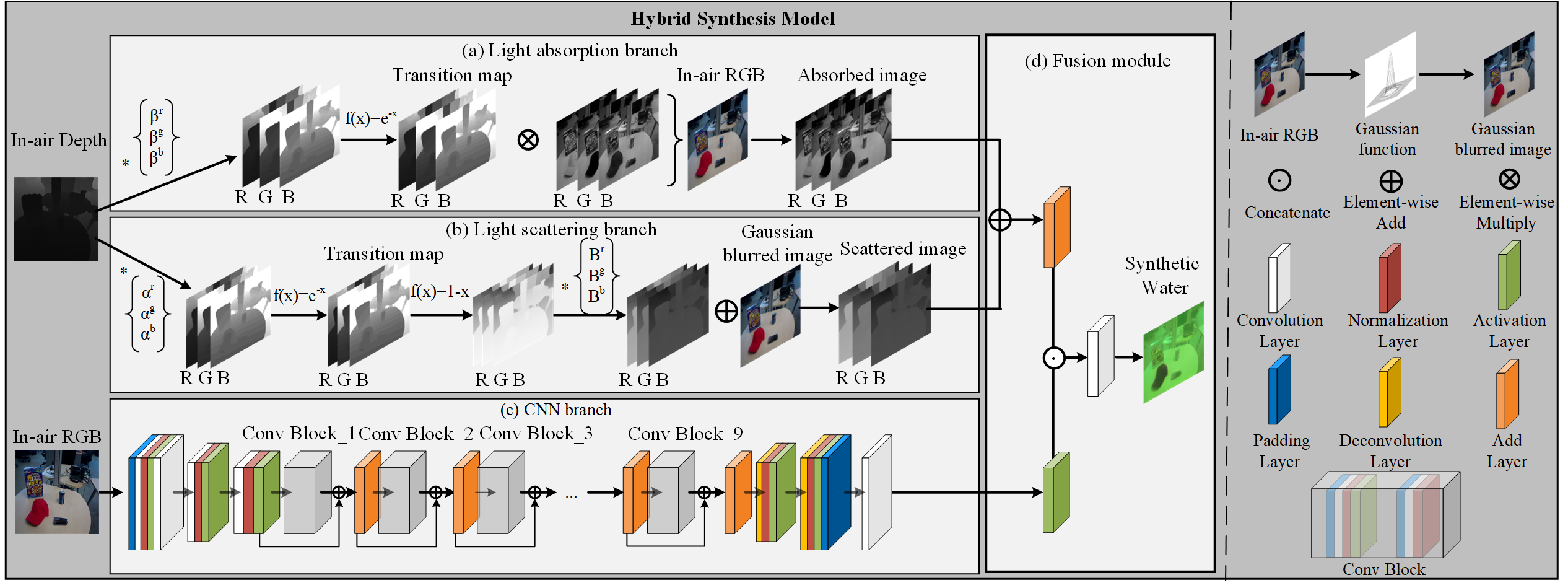}
\caption{The overview of the proposed hybrid synthesis model, which consists of (a) light absorption branch, (b) light scattering branch and (c) CNN branch. The outputs of three branches are combined into the final synthetic underwater image in (c) fusion module.}
\label{fig:AFinal}
\end{figure*}
\subsection{Perceptual loss}
A number of recent papers have employed perceptual loss to improve image quality. One of the means is to use feature reconstruction-based perceptual loss \cite{b33, b34, b35, b36}, which has achieved large successes in a wide variety of image translation tasks, such as transfer learning, image super resolution and synthesis. These works first extract convolutional features from pre-trained CNNs, and then design a perceptual loss function for the other image transformation network, which minimises the discrepancy between the extracted CNN features of the generated and target images which guides the transformation network to generate high-quality images similar to the target images.

The other means is to use classification-based perceptual loss \cite{b37, b38, b31, b32}, which generates images by computing and feeding back the gradients of a given class with respect to the input image in a learnt classifiation network. For example, given a learnt classification network on the large-scale ImageNet dataset and a class of interest, Karen et al. \cite{b28} gave an zero image to the well-trained classification network, then computed the gradients of the given class with respect to the zero input image, and feeded the gradients to update the input image, and finally a newer version of the input image is generated. Since the new image was generated by maximising the score of the given class, some discriminative attributes of the interested object class are visualized in the input images and help to recognise the objects of the interested class. Similar optimisation techniques can also be used to generate high-confidence fooling images \cite{b31, b32}.

The success of perceptual loss based strategies lies in that a high-capacity neural network trained beforehand could implicitly learn to encode relevant image details and semantics. The use of perceptual loss functions allows the transfer of knowledge from the high-capacity neural networks to the transformation networks or the input. However, all these perceptual losses aim to generate high quality images without explicitly considering the requirements of assisting high-level tasks conducted later on. The high capacity networks are trained on image-level labels with a classification loss without detailed object information. In this paper, we propose two detection perceptual losses, guiding the enhancement network to generate visually pleasing or detection-favouring images.
\section{Proposed methods}
\label{sec:proposedmethod}
In this section, we first present the overview of our proposed HybridDetectionGAN framework. Then, we describe the proposed hybrid synthesis model and perceptual enhancement models. Finally, we introduce the training of the framework.

\subsection{The overview of the proposed framework}
As shown in Fig.~\ref{fig:HybridDetectionGAN}, our hybrid synthesis model, enhancement model and detection perceptor are incorporated into a unified Cycle-Consistent Adversarial Networks \cite{b17}, named HybridDetectionGAN. The enhancement model and the detection perceptor construct the complete detection-perceptual enhancement model. The HybridDetectionGAN framework exploits two cycle consistency paths to learn the transformation between underwater and in-air domains from the unpaired images. Particularly, the forward cycle-consistency path starts with real in-air RGB-D images and finishes with reconstructed in-air images (i.e., enhanced underwater images and depth images). The hybrid synthesis model transforms in-air RGB-D images into the underwater counterparts. After sufficient adversarial training, the discriminator has difficulty in distinguishing actual underwater images from the synthetic ones, and the underwater images synthesized by the hybrid synthesis model are deemed closely similar to their real version. The enhancement model transforms underwater images into in-air images, where the cyclic loss (cycle-consistency loss) regularises both the synthesis and enhancement models to generate better structural content in the images. Following the enhancement model, a novel detection perceptor is activated, which is trained with the detection loss on the real in-air images. Then, during the training of the enhancement model, detection-favouring information is given to the enhancement model via the object-focused perceptual loss. The backward cycle-consistency path starts with real underwater images and finishes with the reconstructed underwater ones. The adversarial process between the enhancement model and the discriminator enables more realistic in-air images.

\subsection{Hybrid underwater image synthesis model}
The generalisation of the enhancement model highly relies on the quality of the synthetic training data. To develop a robust synthesis model, we incorporate an improved physical model and a data-driven CNN model into a hybrid synthesis model. The improved physical model is able to simulate evident haze-effects and maintains a coarse view of underwater images by applying the priors of light absorption and scattering. The CNN model works as a supplement to generate finer details of underwater images by modelling other factors that result in the degradation of images. Fig.~\ref{fig:AFinal} shows the overall structure of the hybrid synthesis model. It consists of three branches, i.e., light absorption, light scattering, and CNN branches. The light absorption and scattering branches are built on optical priors, and construct the complete physical model.
\subsubsection{Improved underwater image formation model}
Our underwater images formation model is based on the physical model shown in Eq.~(\ref{eq:oriphysical}) and formulated as Eq.~(\ref{eq:newphysical}).
\begin{equation}
\begin{aligned}
I_{sw}^{\lambda}(x, y) =& \sum_{w=0}^{W}\sum_{h=0}^{H}\sum_{m=0}^{M}I_{con}(x+w, y+h, m)\theta_f^{\lambda}(w,h,m),\\
&\lambda \in \{r,g,b\}
\label{eq:newphysical}
\end{aligned}
\end{equation}
where
\begin{equation}
I_{con} = I_{ab}+ I_{sc} \odot I_{cnn}
\label{eq:concat}
\end{equation}
$I_{sw}(x,y)$ denotes the pixel value of the synthetic underwater image at each point $(x, y)$. $\lambda$ denotes different channels of an image, including red, green, and blue channels. $\theta_f^{\lambda}$ is a $W \times H \times M$ convolutional filter at stride 1, responsible for converting the outputs of the three branches into the $\lambda$-channel of the synthetic underwater image. $I_{ab}$ and $I_{sc}$ indicate the images suffering from light absorption and  scattering respectively, $I_{cnn}$ is the output of the CNN branch. $+$ and $\odot$ denote the element-wise adding operation and channel-wise concatenation operation respectively. $I_{con}$ is the fused output of three branches and, $I_{sw}$ is achieved through the convolution operation between $\theta_f$ and $I_{con}$ in Eq.~(\ref{eq:newphysical}).

Light absorption causes the change of color tones, and the image suffering from light absorption can be described by Eq. (\ref{eq:absorption}). Different channels of a RGB image have different absorption coefficient $\beta$. The effect of light absorption becomes stronger with increasing object-camera distance $d$ as more energy is absorbed by the water. For each pixel on the image, its depth $d$ comes from depth map $I_{d}$. To retain the absorption image, we first compute the transition map $T=e^{-I_{d}\beta^{\lambda}}$, indicating part of the light has been absorbed during the propagation in the water. Then, we compute each channel of the absorption image $I_{ab}^{\lambda}$ using element-wise multiplication operation $\otimes$ between the in-air image $I_{a}$ and $T$.
\begin{equation}\footnotesize
	I_{ab}^{\lambda}=I_{a}^{\lambda}\otimes e^{-I_{d}\beta^{\lambda}}, \lambda \in \{r,g,b\}
\label{eq:absorption}
\end{equation}

Light scattering is an optical process in underwater imaging, including forward and back scattering \cite{b16}. Forward scattering occurs when the light reflected from the object is scattered on its way to the camera, resulting in an effect very similar to the Gaussian blur. However, the commonly used physical model only considers back scattering priors while ignoring forward scattering priors. Hence, we simulate the haze-effects caused by forward scatters using the Gaussian blur function. The light scattered image $I_{sc}$ is formulated as
\begin{equation}
	I_{sc}^{\lambda}=I_{bsc}^{\lambda}+I_{fsc}^{\lambda}, \lambda \in \{r,g,b\}
\label{eq:Isc}
\end{equation}
where
\begin{equation}
	I_{bsc}^{\lambda}=B^{\lambda}(1-e^{-I_{d}\alpha^{\lambda}}), \lambda \in \{r,g,b\}
\label{eq:Ibsc}
\end{equation}

\begin{equation}
	I_{fsc}=I_{a}\Phi(x,y)
\label{eq:Ifsc}
\end{equation}
\begin{equation}
	\Phi(x,y)=Ae^{-\frac{x^2+y^2}{q^2}}
\label{eq:gaussian}
\end{equation}
where A is determined by
\begin{equation}
	\int\int Ae^{-\frac{x^2+y^2}{q^2}}dxdy=1
\label{eq:constrains}
\end{equation}
where $I_{bsc}$ and $I_{fsc}$ are the images suffering from backward and forward scattering, respectively. $B$ and $\alpha$ denote the global background light and the backscatter coefficient. $\Phi(x,y)$ denotes the Gaussian function with the kernel size 5, $q$ is the scale, $q=5 \sim 7$ in our experiments.

Different from the previous physical model-based synthesis methods, which use predefined parameters to synthesize 10 Jerlov water types, the parameters of our proposed underwater image formation model can be learnt from big data using gradient-based optimisation algorithms and better simulate the characteristics of the target underwater images. Moreover, a CNN branch works as a supplement of the physical model to simulate more degraded characteristics of underwater images. The light absorption and scattering priors of the physical model can simulate coarse color distortion and haze-effects. However, significant errors exist in the resultant images since other factors also affect the formation of the underwater images. For instance, the existence of artificial lights leads to non-uniform illumination on the images, and the movement of the cameras bring in noise in the captured images. All of these factors have not been modelled in the physical model, and the CNN branch helps us to simulate these factors and generates finer color tones, illumination change and noise. The detailed structure of the CNN branch can be found in Fig. \ref{fig:AFinal}.

\subsection{Detection perceptual enhancement model}
The perceptual loss have been widely used to improve the quality of images or generate images of interests, and its success in many tasks demonstrates that a high-capacity CNN trained for a high-level task has the capability of implicitly encoding fine image details and task-related semantic knowledge. Inspired by these observations, we first train a one-stage deep detector \cite{b6} on high quality in-air images, and it encodes object details and potential detection favouring information of high quality in-air images. Then, we add this deep detector as a detection perceptor (its weights are kept fixed to those obtained from training) to provide detection-favouring perceptual information to the enhancement model. 

During the adversarial training, the enhanced image is forwarded to the detection perceptor. The detection perceptor is an one-stage deep detector \cite{b32} as shown in Fig.~\ref{fig:HybridDetectionGAN}, which first associates 6 default patches of different scales and aspect ratios at each location of 7 convolutional layers (we only show 2 default patches on the $9$-th convolutional layer in Supplementary A Fig.~\ref{fig:dpenhancement} for simplicity). Then, the predictor works with a $C+1$-dimension class vector and a $4$-dimension location vector for each default patch using a $3 \times 3$ convolutional filter. $C$ denotes the number of the object classes and 1 denotes the background class. Next, it assigns the ground-truth class and location for each default patch using the following matching rules: If the Interaction over Union (IoU) between a default patch and its overlapped ground-truth object is larger than 0.5, the class and the location of the object are assigned as those of the default patch. If the default patch does not match any ground-truth object with an IoU larger than 0.5, it is labelled as the background patch and has no ground-truth location. For example, the default patch (red) shown in Fig.~\ref{fig:dpenhancement} matches the cereal box (green box) with the IoU larger than 0.5, so it is an object patch and its ground-truth class and location are labelled as those of the cereal box. The black patch does not match any object, so it is labelled as the background patch and has no target location. Finally, a detection based perceptual loss computes the discrepancies between patches on the enhanced image and that on the high quality in-air image, and feeds back the discrepancies to the enhancement model in the form of gradients, based on which the enhancement model continuously updates its parameters. Until no gradient has any impact, indicating the enhanced images are the same as the in-air images in the detection perceptor space, i.e., the object details and detection favouring information of the in-air image encoded in the detection perceptor space have been properly transformed to the enhancement model.

We design two perceptual loss functions having different objectives for two detection perceptors. The first loss is named patch detection perceptual loss that aims to generate  patch-level realistic in-air images. The second one is named object-focused detection perceptual loss which aims to generate detection-favouring images that can improve the detection accuracy. Fig.~\ref{fig:dpenhancement} shows the overall structure of the patch detection perceptual enhancement model and the object-focused detection perceptual enhancement model, each of which consists of an enhancement model and a detection perceptor with a specially designed perceptual loss.
\subsubsection{Patch detection based perceptual loss}
Patch detection perceptual loss function $L_{p}$ is an one-stage detection loss \cite{b32}, which is a weighted sum of classification loss $L_{cls}$ and localization loss $L_{loc}$.
\begin{equation}\small
\begin{split}
	L_{p} =&\frac{1}{N}\sum_{i \in all} L_{cls}(pcls^i, gcls^i)+ \frac{1}{\bar{N}}\sum_{i \notin bg}L_{loc}(ploc^i, gloc^i)
\end{split}
\end{equation}
where $pcls^i$ and $gcls^i$ denote the predicted and ground-truth class vectors of $C+1$ dimensions for the $i$-th default patch. $ploc^i$ and $gloc^i$ denote the predicted and ground-truth location vector of $4$-dimensions for the $i$-th default patch. The $4$-dimension location vector includes the coordinates of center $(cx, cy)$ with width $w$ and height $h$. $all$ and $bg$ are the set of all the default patches and the patches belonging to the background samples, which do not contribute to the location loss because of no ground-truth location. $N$ and $\bar{N}$ are the numbers of all the default and the object patches. Specially, the classification loss $L_{cls}$ is a Softmax loss.
\begin{equation}
\begin{aligned}
	L_{cls}(pcls, gcls) = -\sum_{c=1}^{C+1}pcls_{c}log(gcls_{c})
\end{aligned}
\end{equation}
where $pre\_cls_{c}$ and $gt\_cls_{c}$ indicate the $c$-th element of the predicted and the ground-truth class vectors, respectively. The localisation loss is a Smooth L1 loss \cite{b41} between the predicted and the ground-truth locations.
\begin{equation}
\begin{aligned}
	L_{loc}(ploc, gloc) = \sum_{l=1}^{4}smooth_{L1}(ploc_{l}-gloc_{l})
\end{aligned}
\end{equation}
$L_{cls}$ encourages the enhancement model to generate images which minimise the class discrepancy between the generated and the ground-truth patches, and $L_{loc}$ encourages the enhancement model to generate images of minimising the location discrepancy between the generated and the ground-truth patches, thus the patch perceptual detection loss can guide the enhancement model towards generating more realistic patches at more accurate locations.

\subsubsection{Object-focused detection based perception loss}
For underwater object detection applications, many objects look very similar to the background. The complex background may degrade the detection accuracy of the detectors. To deal with this challenge, we propose the object-focused detection based perceptual loss $L_{of}$:
\begin{equation}
\begin{split}
L_{of} =& \frac{1}{\bar{N}} \sum_{i\notin bg} L_{cls}(pcls^i, gcls^i) + L_{loc}(ploc^i, gloc^i)
\end{split}
\label{eq:ofperceptual}
\end{equation}
Different from the patch detection perceptual loss, the object-focused detection perceptual loss only focus on feeding back the informations of object patches, while ignoring the background patches, as shown in Fig.~\ref{fig:dpenhancement}, the black cross indicates background patches have no information feedback to the enhancement model ). From the optimisation perspective, $L_{of}$ is designed towards assigning ground-truth classes and locations of the object patches on the enhanced image, that equals to improving the detection accuracy of the deep detector trained on the enhanced images.

\subsection{Training of our overall HybridDetectionGAN}
We first train the standard one-stage deep detector on high quality in-air images. Afterwards, we add the detector after the enhancement model in the forward cycle-consistency path. We then move on to train the synthesis and enhancement models.

\subsubsection{Training of the hybrid synthesis model}
Denoting $G_{\theta_{a2w}}$ as the hybrid synthesis model parameterised by $\theta_{a2w}$ and $\theta_{cnn}$ as the parameters of the CNN branch, then we have $\theta_{a2w}=\{\alpha, \beta, B, \theta_{cnn}, \theta_f\}$. Denote $G_{\theta_{w2a}}$ as the enhancement model parameterised by $\theta_{w2a}$. We obtain $\theta_{a2w}$ by minimising the loss function $L_{a2w}$, which is a combination of an adversarial loss $L_{adv\_w}$ and a cycle-consistency loss $L_{cyc\_w}$.
\begin{equation}
\begin{aligned}
L_{a2w}=w_{1}L_{adv\_w}+w_{2}L_{cyc\_w}
\end{aligned}
\label{eq:La2w}
\end{equation}
The first term is an adversarial loss produced by the discriminator Water, which is denoted as $D_{\theta_{dw}}$ and parameterised by $\theta_{dw}$. Taking the synthetic underwater image $G_{\theta_{a2w}}(I_{a},I_{d})$ as the input, the discriminator outputs the estimated probability of the synthetic underwater image treated as a real underwater image, denoted as $D_{\theta_{dw}}(G_{\theta_{a2w}}(I_{a}, I_{d}))$. By fooling the discriminator with the synthetic underwater image, the adversarial loss is formulated as $L_{adv\_w}=-log D_{\theta_{dw}}(G_{\theta_{a2w}}(I_{a}, I_{d}))$, which encourages the hybrid synthesis model to produce more realistic underwater images. The cycle-consistency loss $L_{cyc\_w}$ is computed as the $L_1$ distance between the reconstructed and ground-truth underwater image $I_{w}$, i.e., $L_{cyc\_w}=||G_{\theta_{a2w}}(G_{\theta_{w2a}}(I_{w}))-I_{w}||_{1}$.

Different from the physical-model based synthesis  model, our physical parameters $\alpha$, $\beta$ and $B$ can better simulate the characteristics of the target underwater images since they are learnt from the training data via the gradient descent optimisation algorithm. The optimisation algorithm iteratively updates the parameter $\beta^{\lambda}$ by
\begin{equation}
\begin{split}
\beta^{\lambda}=\beta^{\lambda}-\eta\frac{\partial L_{a2w}}{\partial \beta^{\lambda}}
\end{split}
\label{eq:19}
\end{equation}
where $\eta$ is the learning rate. In order to update $\beta^{\lambda}$, we need to compute $\frac{\partial L_{a2w}}{\partial \beta}$, which indicates the gradient of $L_{a2w}$ with respect to $\beta^{\lambda}$. Denote $I_{add}^{\lambda}$ as the output of the physical branch, $I_{add}^{\lambda}=I_{ab}^{\lambda}+I_{sc}^{\lambda}$. $I_{ab}^{\lambda}$, $I_{con}$ and $I_{sw}$ are formulated in Eqs.~(\ref{eq:newphysical})-(\ref{eq:absorption}). We can derive $\frac{\partial L_{a2w}}{\partial \beta}$ from the chain rule in Supplementary B and finally achieve $\beta^{\lambda}$ by Eq.~\ref{eq:20}.
\begin{equation}
\begin{aligned}
\beta^{\lambda}=&\beta^{\lambda}+\eta(w_1\frac{\partial L_{adv\_w}}{\partial I_{sw}}+w_2\frac{\partial L_{cyc\_w}}{\partial I_{sw}})\\
&(I_{ab}^{\lambda}\otimes e^{-I_d \beta^{\lambda}}I_d \frac{\partial I_{sw}}{\partial I_{con}} \frac{\partial I_{con}}{\partial I_{add}^{\lambda}})
\end{aligned}
\label{eq:20}
\end{equation}
Similarly, we update $\alpha$ and $B$ through the chain rule, and achieve the final parameters when the models converge.
\begin{equation}
\begin{aligned}
\alpha^{\lambda}=&\alpha^{\lambda}-\eta(w_1\frac{\partial L_{adv\_w}}{\partial I_{sw}}+w_2\frac{\partial L_{cyc\_w}}{\partial I_{sw}})\\
&(B^{\lambda} e^{-I_d \alpha^{\lambda}}I_d \frac{\partial I_{sw}}{\partial I_{con}} \frac{\partial I_{con}}{\partial I_{add}^{\lambda}})
\end{aligned}
\end{equation}
\begin{equation}
\begin{aligned}
B^{\lambda}=&B^{\lambda}-\eta(w_1\frac{\partial L_{adv\_w}}{\partial I_{sw}}+w_2\frac{\partial L_{cyc\_w}}{\partial I_{sw}})\\
&((1-e^{-I_d \alpha^{\lambda}}) \frac{\partial I_{sw}}{\partial I_{con}} \frac{\partial I_{con}}{\partial I_{add}^{\lambda}})
\end{aligned}
\end{equation}
We obtain $\theta_{dw}$ by optimising the loss function $L_{d\_w}$, which encourages the discriminator to distinguish the difference between the synthetic and real underwater images:
\begin{equation}\small
\begin{split}
L_{d\_w}=-logD_{\theta_{dw}}(I_{w})-log(1-D_{\theta_{dw}}(G_{\theta_{a2w}}(I_{a}, I_{d})))
\end{split}
\end{equation}

\subsubsection{Training of the detection perceptual enhancement model}
We obtain $\theta_{w2a}$ by optimising the loss function $L_{w2a}$, which is a weighted combination of adversarial loss $L_{adv\_a}$, cycle-consistency loss $L_{cyc\_a}$, and perceptual loss $L_{of}$ (or $L_{patch}$).
\begin{equation}
\begin{split}
L_{w2a}=w_{1}L_{adv\_a}+w_{2}L_{cyc\_a}+w_{3}L_{of}
\end{split}
\end{equation}

Denote $D_{\theta_{da}}$ as the discriminator parameterised by $\theta_{da}$. Taking the enhanced RGB-D underwater image $G_{\theta_{w2a}}(I_{w})$ as input, the discriminator outputs the estimated probability of the enhanced image as a real in-air image, denotes as $D_{\theta_{da}}(G_{\theta_{w2a}}(I_{w}))$. To fool the discriminator with the enhanced underwater image, the adversarial loss $L_{adv\_a}$ is formulated as $L_{adv\_a}=-log D_{\theta_{da}}(G_{\theta_{w2a}}(I_{w}))$. $L_{cyc\_a}$ is computed as the $L_1$ distance between the reconstructed and ground-truth in-air image, $
L_{cyc\_a}=||G_{\theta_{w2a}}(G_{\theta_{a2w}}(I_{a}))-I_{a}||_{1}$. $L_{of}$ is the object-focused detection perceptual loss that encourages the enhancement model to generate detection-favouring outcomes.

We obtain $\theta_{da}$ by optimising $L_{d\_a}$, which encourages the discriminator to distinguish the difference between the enhanced underwater images and the real in-air images.
\begin{equation}
\begin{split}
L_{d\_a}=-logD_{\theta_{da}}(I_{a})-log(1-D_{\theta_{da}}(G_{\theta_{w2a}}(I_{w})))
\end{split}
\end{equation}
\subsubsection{How the detection perceptor influence the enhancement model in the form of gradients}
During the training of the enhancement model, the optimisation algorithm iteratively updates the enhancement model's parameter $\theta_{w2a}$ by
\begin{equation}
\begin{aligned}
\theta_{w2a}&=\theta_{w2a}-\eta\frac{\partial L_{w2a}}{\partial \theta_{w2a}}\\
&=\theta_{w2a}-\eta(w_1\frac{\partial L_{adv\_a}}{\partial \theta_{w2a}}+w_2\frac{\partial L_{cyc\_a}}{\partial \theta_{w2a}}+w_3\frac{\partial L_{of}}{\partial \theta_{w2a}})\\
\end{aligned}
\label{eq:of1}
\end{equation}
\begin{equation}
\begin{aligned}
\frac{\partial L_{of}}{\partial \theta_{w2a}}=\frac{1}{\bar{N}}\sum_{i\notin bg} \frac{\partial L_{cls}(pcls^i, gcls^i)}{\partial \theta_{w2a}}+\frac{\partial L_{loc}(ploc^i, gloc^i)}{\partial \theta_{w2a}}
\end{aligned}
\label{eq:of}
\end{equation}
In each iteration, the detection perceptor feeds back the gradients $\eta w_3\frac{\partial L_{of}}{\partial \theta_{w2a}}$ to the enhancement model. From Eqs.~(\ref{eq:of1}) and (\ref{eq:of}), we can see that the enhancement model continuously updates its parameter $\theta_{w2a}$ to minimise $L_{cls}(pcls^i, gcls^i)$ and $L_{loc}(ploc^i, gloc^i)$, equivalently maximising the class prediction accuracy and location prediction accuracy of object patches. Thus the gradients $\eta w_3\frac{\partial L_{of}}{\partial \theta_{w2a}}$ help the enhancement model to generate images enabling more accurate object detection in the following process.

\section{Experimental Setup}
\label{sec:exmperiments}
To demonstrate the effectiveness of the proposed method, we conduct comprehensive evaluations on both the unpaired ChinaMM-MultiView dataset and paired OUC dataset. In this section, we first introduce the experimental datasets and evaluation metrics. Then, we describe the implementation details.
\subsection{Datasets}
\textbf{The unpaired ChinaMM-MultiView dataset} is constructed by collecting images from an underwater image dataset ChinaMM \cite{b1} and an in-air image dataset MultiView \cite{b39}. ChinaMM is a public competition dataset for evaluating UIE algorithms. The dataset has publicly released the train set of 2,071 images and the validation set of 676 images. This dataset provides bounding box annotations and contains three object categories: seacucumber, seaurchin and scollap. The resolution of each image is 720x405 pixels. The in-air dataset Multiview  consists of 14,179 training images and 1,206 testing images which are captured in in-door scenes with high image quality. This dataset provides RGB images (640$\times$480 pixels), depth images and bounding box annotations. It contains five object categories: bowl, cap, coffee mug, cereal box and soda can. To construct the unpaired ChinaMM-MultiView dataset, we randomly choose 2,071 images as the training set of MultiView and 676 images as the testing set of MultiView.

\textbf{The paired OUC dataset} \cite{b40} provides underwater images, high quality reference images and bounding box annotations. The training set contains 2,597 image pairs where the testing set contains 1,198 image pairs. The dataset does not provide depth images which are needed by our hybrid synthesis model, so we apply the technology reported in \cite{b8} to obtaining depth maps for all the reference images. 

\subsection{Evaluation Metrics}
We conduct extensive experiments to quantitatively and qualitatively evaluate the proposed hybrid synthesis model and detection perceptual model. For the qualitative evaluations, we directly present the resultant images. For the quantitative evaluations, we apply several commonly used full-reference image quality evaluation metrics, where ground-truth or references are available. The full-reference metrics include the widely used Mean Square Error (MSE), Structural Similarity (SSIM) and Peak Signal-to-Noise Ratio (PSNR), and the Patch-based Contrast Quality Index (PCQI) \cite{b41}. For the experiments with no reference image, we apply two non-reference image quality evaluation metrics, including underwater image quality measure (UIQM) \cite{b42} and underwater color image quality evaluation (UCIQE) \cite{b43}. UIQM is a linear combination of three components, i.e., the Underwater Image Colorfulness Measure (UICM), Underwater Image Sharpness Measure (UISM), and Underwater Image Contrast Measure (UIConM). In some experiments, we also give the values of these three components for detailed discussion. In addition to the image quality evaluation metrics, we train the deep detector \cite{b6} using the enhanced images by different UIE algorithms and use the mean Average Precision (mAP) as a detection task-specific evaluation metric to evaluate different UIE algorithms.

\subsection{Implementation details}
All the experiments are conducted on a server with an Intel Xeon CPU @ 2.40GHz and 2 parallel Nvidia Tesla P100 GPUS. We implement the proposed HybridDetectionGAN framework using the Keras framework. We train the detection perceptor using the Adam optimiser \cite{44} with 120 epochs and an initial learning rate of 1e-3. The learning rate is decreased by a factor of 10 after 60 epochs. We train HybridDetectionGAN for 200 epochs. The initial learning rate of the hybrid synthesis model, the perceptual enhancement model and two discriminators are 2e-4, and after 100 epochs, we apply a linear decay of the learning rate for all four components. The source code will be made available at:https://github.com/LongChenCV/HybridDetectionGAN.

\section{Results and discussion}
\label{sec:exmperimentalresults}
In this section, we present and discuss the experimental results and findings. We first conduct ablation experiments to investigate the influence of different components on our proposed HybridDetectionGAN framework. Then, we compare our method against several state-of-the-art methods on  the two datasets. Finally, we investigate how these UIE algorithms influence the deep detectors in the following process.

\subsection{Ablation studies}
The proposed HybridDetectionGAN framework integrates a hybrid underwater image synthesis model and a detection perceptual underwater enhancement model. We conduct ablation experiments in order to evaluate them on both unpaired ChinaMM-MultiView and paired OUC datasets.

\subsubsection{\textbf{Ablation studies of the hybrid synthesis model}}
We conduct ablation experiments to investigate how the CNN, scattering and absorption branches and the complete physical branch (consisting of the scattering and absorption branches) influence the synthetic results.

\begin{figure}[htb]
\centering
\includegraphics[height=4.5cm, width=9cm]{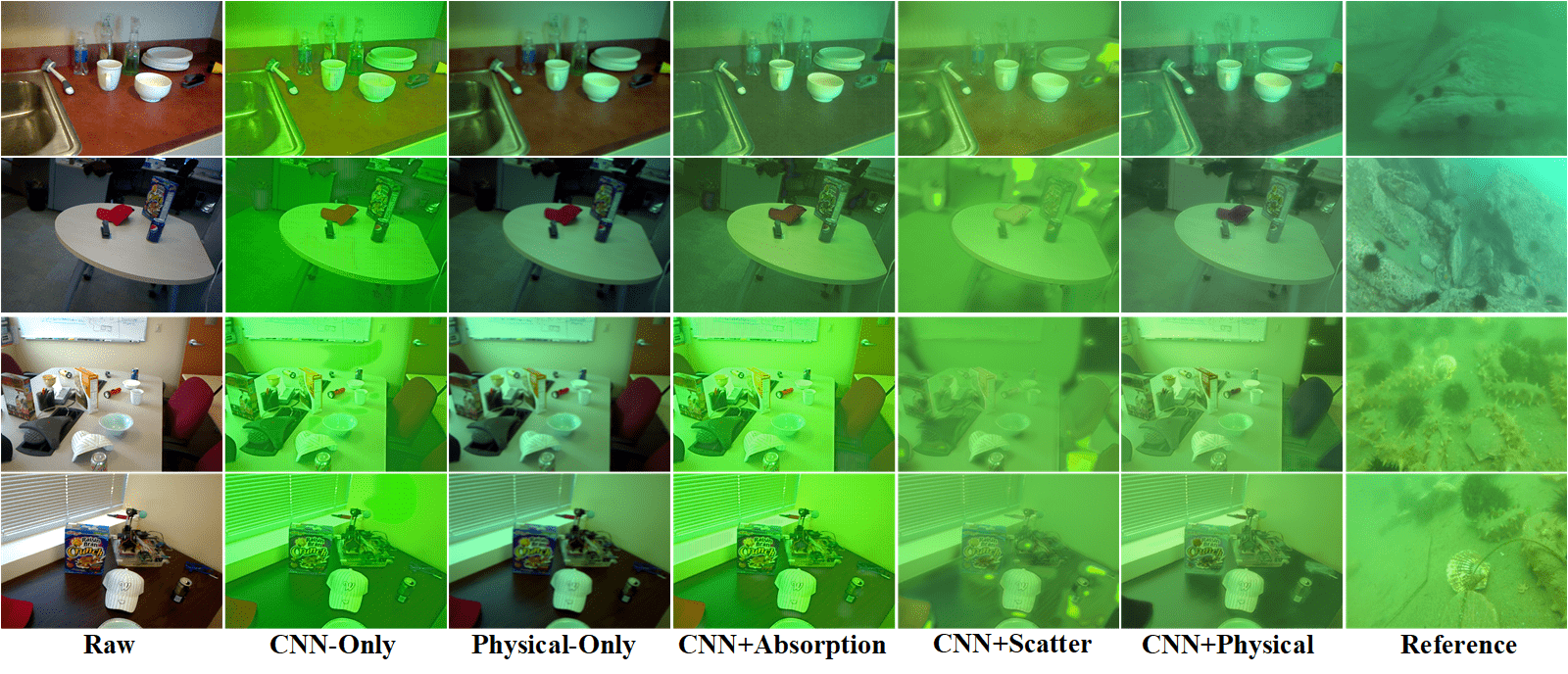}
\caption{Qualitative comparison of synthesis models with different component settings on MultiView. From left to right are raw in-air images, the results of the synthesis models with only the CNN branch, only the physical branch, CNN and absorption branches,  and CNN and scattering branches, complete hybrid synthesis model, and reference underwater images from ChinaMM. Best viewed in the digital form.}
\label{fig:ablation_syn_Multiview}
\end{figure}
Fig.~\ref{fig:ablation_syn_Multiview} presents the qualitative comparison of the synthesis models with different component settings on the MultiView dataset. We observe that our complete hybrid synthesis model produces the visual appearance most similar to the real underwater images of ChinaMM. The synthesis models without the scattering branch have a limited capability of modelling the haze-effects. For example, the resultant images of CNN-Only and CNN+Absorption synthesis models are relatively clear in spite of severe color distortion. After having incorporated the scattering prior into these two models, we clearly witness the haze-effects on the images. Both CNN-Only and Physical-Only synthesis models do not generate diverse results. This is because the former one usually runs into the classical mode collapse problem that only produces outcomes of a single mode, e.g., all the synthetic images are of the same color tone. The latter also generates underwater images in a monotonous style. Once the physical model has been trained, only one fixed set of parameters have been optimised, leading to optimal results within a specific environment. When we integrate the CNN and absorption branches, diverse results are obtained. The absorption prior helps CNN to generate images with different color tones whilst avoiding the artifact problem. For example, artifacts frequently occur in the resultant images of the synthesis models without the absorption branch, such as the CNN-Only and CNN+scattering synthesis models. Supplementary C presents the qualitative comparison of the synthesis models with different component settings on the OUC dataset.

\begin{table}[h]
\begin{center}
\renewcommand\tabcolsep{3.5pt}
\caption{Quantitative comparison of the synthesis models with different components on the OUC dataset.}
\label{table:quantiSyn_OUC_component}
\begin{tabular}{l|ccc|llll}
\hline
Model & CNN & Absorption & Scatter & MSE & PSNR & SSIM & PCQI\\
\hline
\multirowcell{5}{Synthesis\\Models} & \checkmark & & & 0.2777 & 23.7386 & 0.7664 & 0.6627\\
& \checkmark & \checkmark & & 0.1360 & 26.8716 & 0.8963 & 0.9398\\
& \checkmark & & \checkmark & 0.1289 & 27.0851 & 0.9046 & 0.9364\\
& & \checkmark & \checkmark & 0.2882 & 23.5787 & 0.7659 & 0.6600\\
& \checkmark & \checkmark & \checkmark & \textbf{0.0978} & \textbf{28.2895} & \textbf{0.9131} & \textbf{0.9518}\\
\hline
\end{tabular}
\end{center}
\end{table}
In addition to the qualitative comparison, we also use four full-reference image quality evaluation metrics to evaluate the synthesis models supported by the reference images in the OUC dataset. From Table~\ref{table:quantiSyn_OUC_component}, we observe the superiority of the complete hybrid synthesis model over the other models as to four metrics. This indicates the synthetic underwater images of the hybrid model are the closest ones to the reference images. After having removed the absorption branch, the values of the four metrics decrease due to the existence of frequent artifacts. Removing the scattering branch also decreases the values of the four metrics due to the haze-effects. The synthesis model with the only physical model generates images with color distortion and the worst quantitative scores.

\subsubsection{\textbf{Ablation studies of the enhancement model}}
We first investigate the influence of the quality of the training data, i.e., the synthetic underwater images, on the enhancement model. Then, we analyse how the detection perceptor affects the enhancement model.\\
\textbf{The influence of the quality of the synthetic underwater images on the enhancement model.} We divide the synthetic underwater images into four categories: (A) Synthetic underwater images with incorrect color tones generated by the Physical-Only synthesis model; (B) Synthetic underwater images without evident haze-effects generated by the CNN+absorption synthesis model; (C) Synthetic underwater images with artifacts generated by the CNN+scattering synthesis model; (D) Synthetic images with pleasing appearance generated by the hybrid synthesis model.

\begin{figure}[htb]
\centering
\includegraphics[height=4.0cm, width=9cm]{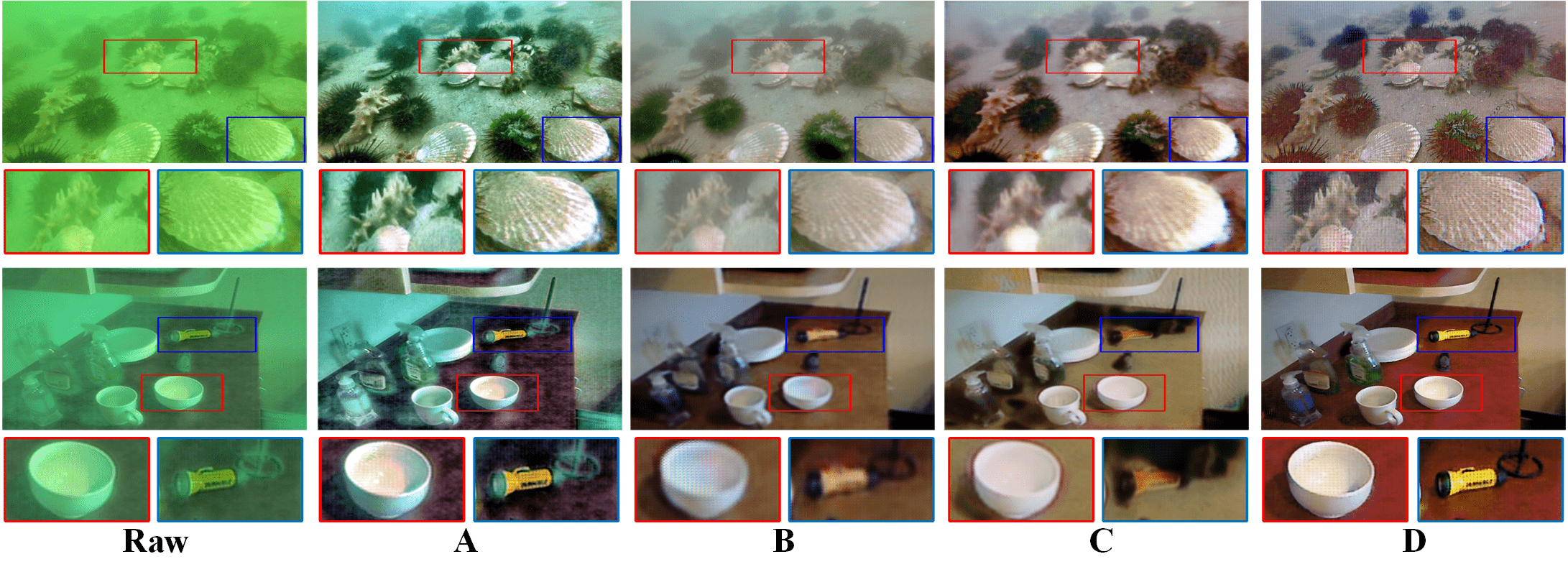}
\caption{Qualitative comparison of the enhancement models trained on different synthetic underwater images on ChinaMM (top row) and MultiviewUnderwater (bottom row). From left to right are the raw underwater images, the results of the enhancement model trained on (A) synthetic images with incorrect color tones, (B) synthetic images without evident haze-effects, (C) synthetic images with artifacts, and (D) synthetic images with pleasing appearance.}
\label{fig:ablation_enhance_two}
\end{figure}
For the unpaired ChinaMM-MultiView dataset, we test different enhancement models on two sub-datasets, i.e., the synthetic underwater dataset, MultiviewUnderwater, which is generated by our hybrid synthesis model with the RGB-D in-air images of MultiView, and the real-world underwater dataset, ChinaMM. Fig.~\ref{fig:ablation_enhance_two} shows the qualitative comparison of the enhancement models trained on different synthetic underwater images on ChinaMM and MultiviewUnderwater. We observe that the enhancement model trained on (A) (synthetic images with incorrect color tones) cannot correct the color casts due to the lack of learning on color transformation between underwater and in-air images. The enhancement model trained on (B) (haze-free synthetic underwater images) leaves evident haze-effects on its results while the enhancement model trained on (C) (synthetic underwater images with artifacts) further aggravates the artifacts problem in the resultant images. In contrast, the enhancement model trained on (D) (visually pleasing synthetic images) performs much better in removing haze-effects and artifacts. However, the results of the enhancement model trained on (D) still suffer from minor artifacts and haze-effects. This is because the deep enhancement model is constructed using fully convolutional layers, which have a limited ability to remove the artifacts and haze-effects.
\begin{table}[htb]\scriptsize
\begin{center}
\renewcommand\tabcolsep{2pt}
\caption{Quantitative comparison of the enhancement models trained with different synthetic underwater images on the MultiviewUnderwater and ChinaMM datasets.}
\label{table:quantiMVU_syn}
\begin{tabular}{l|ccccc|cccc}
\hline
& \multicolumn{5}{c|}{MultiviewUnderwater} & \multicolumn{3}{c}{ChinaMM}\\
\hline
Enhancement models & MSE & PSNR & SSIM & PCQI & mAP & UCIQE & UIQM & mAP\\
\hline
Model trained on A  & 1.3061 & 15.5623 & 0.1906 & 0.5294 & 75.9 & 23.2144 & 2.3864 & 72.2\\
Model trained on B & 1.0058 & 17.0883 & 0.2971 & 0.5549 & 76.3 & 25.2220 & 2.7138 & 74.0\\
Model trained on C & 0.9249 & 18.4655 & 0.3831 & 0.5425 &  72.1 & 25.6417 & 3.1529 & 71.1\\
Model trained on D & \textbf{0.7153} & \textbf{20.5488} & \textbf{0.5632} & \textbf{0.5677} & \textbf{77.5} & \textbf{27.3206} & \textbf{3.8196} & \textbf{76.5}\\
\hline
GT in-air image & 0.0000 & Inf & 1.0000 & 1.0000 & 79.9 & - & - & -\\
GT underwater image & - & - & - & - & - & 21.2847 & 1.4306 & 68.6\\
\hline
\end{tabular}
\end{center}
\end{table}

Tables~\ref{table:quantiMVU_syn} reports the quantitative scores of the enhancement models trained on different synthetic underwater images on the testing sets of MultiviewUnderwater and ChinaMM. We also list the full-reference scores of the ground-truth in-air images on MultiviewUnderwater and the non-reference scores of the ground-truth underwater images of ChinaMM as references. In terms of image quality evaluation metrics, the enhancement model trained on (D) visually pleasing images performs the best while the one trained on (A) images with color distortion achieves the worst scores. The color cast in the results of the latter enhancement model leads to the decreasing scores of the image quality metrics. In terms of mAP, the model trained on (C) images with artifacts achieves the lowest score even though it has relatively higher quantitative scores of image quality evaluation metrics. The artifacts smear the details of the images or objects, deteriorating the detection accuracy more than the incorrect color casts and haze-effects. The enhancement models trained on the visually pleasing images obtain the best scores of image quality evaluation metrics and mAP on the two datasets. In summary, the enhancement model trained on more realistic synthetic underwater images can learn more accurate mappings between the images in underwater  and in-air domains and be better generalised on the real-world underwater dataset.\\
\textbf{The influence of the detection perceptor on the enhancement model.} We compare three enhancement models with different detection perceptor settings, i.e., enhancement model without detection perceptor (denoted as OursWDP), enhancement model with a patch detection perceptor (denoted as OursPatch), and enhancement model with an object-focused detection perceptor (denoted as OursOF).

\begin{figure}[htb]
\centering
\includegraphics[height=4.3cm, width=9cm]{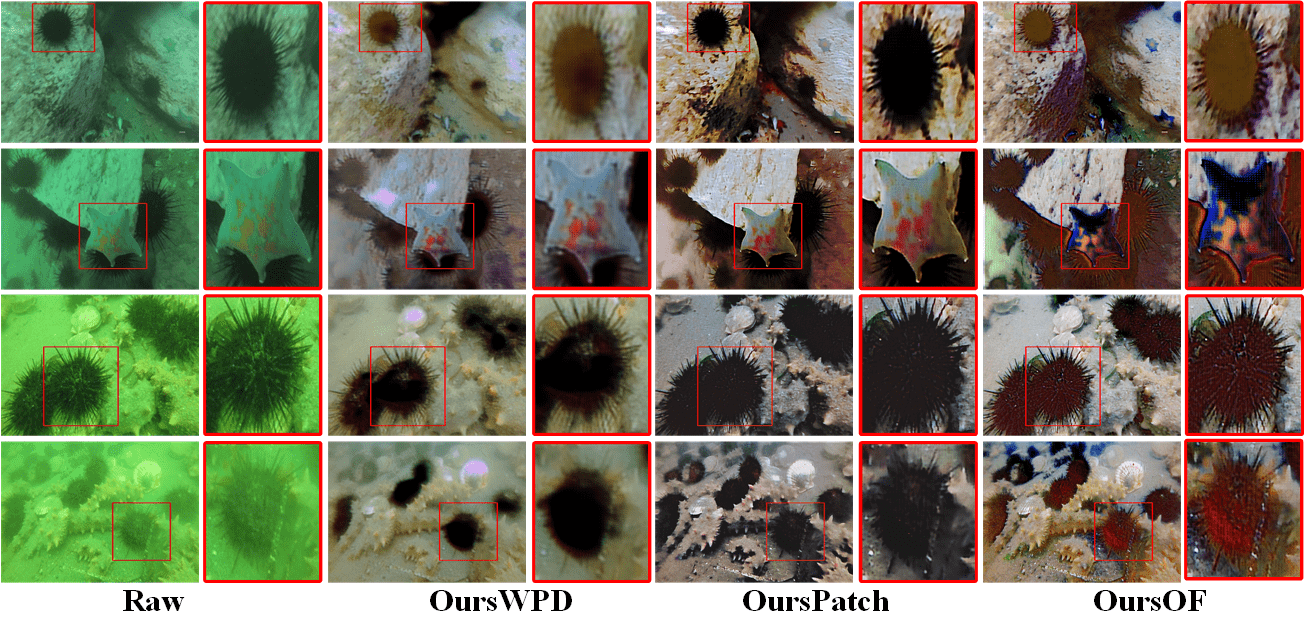}
\caption{Qualitative comparison of the enhancement models with different detection perceptor setting on ChinaMM. From left to right are the raw underwater images, the results of OursWDP, OursPatch, and OursOF, respectively.}
\label{fig:ablation_enhance_ChinaMM_perceptor}
\end{figure}
Fig.~\ref{fig:ablation_enhance_ChinaMM_perceptor} and Supplementary D present the qualitative comparison of the enhancement models with different detection perceptor settings on ChinaMM and MultiviewUnderwater, respectively. We observe that without a detection perceptor, the enhanced results of OursWDP on the two datasets still contain artifacts and haze-effects. OursPatch removes artifacts and restores the details of the image patches such as color tones, visibility, and saturation on the two datasets. This is because the patch detection perceptor trained on high quality in-air images can properly learn potential attributes of high visual quality. These potential attributes, in the form of gradients, help restore the details of image patches. We notice that the detection perceptor is only trained on the in-air images of MultiView, and the object categories of MultiView are different from those of ChinaMM, however, the detection perceptor helps the enhancement model improve the quality of the images of ChinaMM. This indicates that a well-trained detection perceptor not only encodes category-dependent attributes but also category-agnostic attributes such as high quality edges, textures and colors. Compared to OursPatch, OursOF seems to generate sharp objects with high contrast between the objects and the background. In several cases, it also generates over-saturation on the objects, especially for the images of ChinaMM. This may be due to the weak restoration of the background using the object-focused loss function.

\begin{figure}[h]
\centering
\includegraphics[height=6cm, width=9cm]{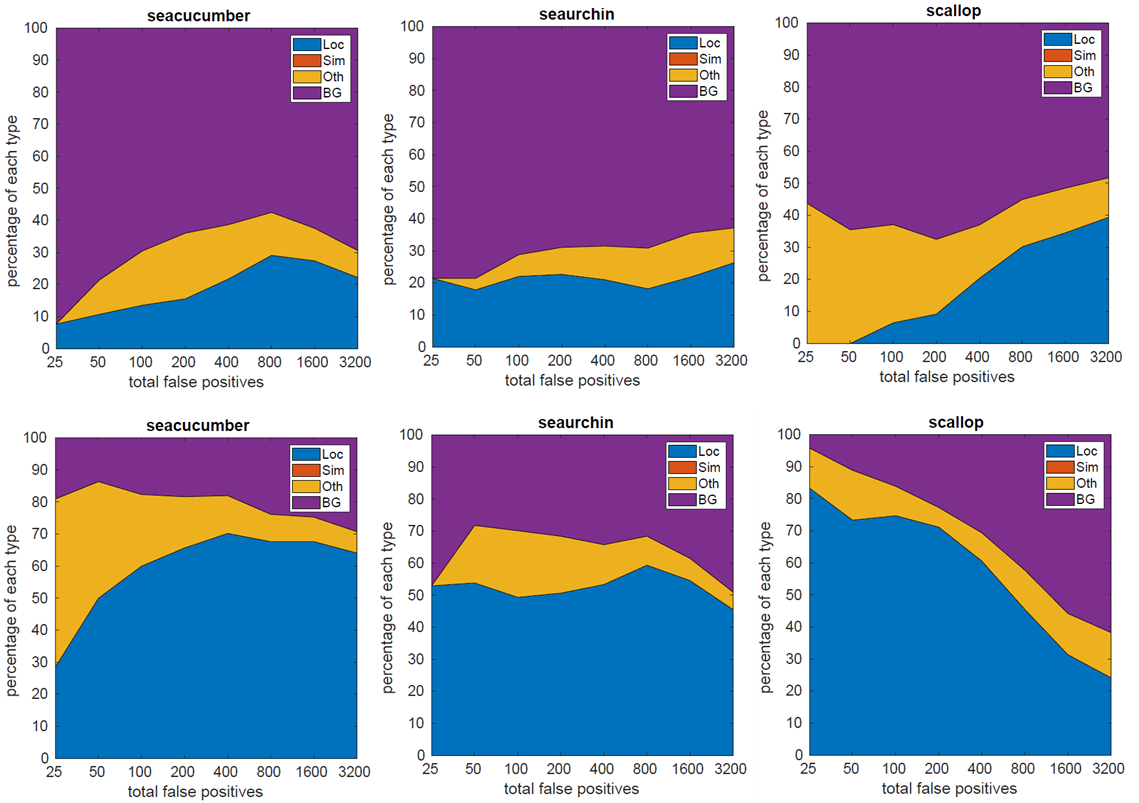}
\caption{The distribution of the top-ranked false positive measures for images of ChinaMM. The false positive measures include localisation errors (Loc), confusion with similar categories (Sim), with others (Oth), or with background (BG). The top row shows the results of OursPatch and the bottom row shows the results of OursOF.}
\label{fig:error}
\end{figure}
\begin{table}[tb]\scriptsize
\begin{center}
\renewcommand\tabcolsep{3.5pt}
\caption{Quantitative comparison of the enhancement models with different detection perceptor settings on the MultiviewUnderwater and ChinaMM datasets.}
\label{table:quantiMVU_det}
\begin{tabular}{l|ccccc|ccc}
\hline
& \multicolumn{5}{c|}{MultiviewUnderwater} & \multicolumn{3}{c}{ChinaMM}\\
\hline
Models & MSE & PSNR & SSIM & PCQI & mAP & UCIQE & UIQM & mAP\\
\hline
OursWDP & 0.7153 & 20.5488 & 0.5632 & 0.5677 & 77.5 & 27.3206 & 3.8196 & 76.5\\
OursPatch & \textbf{0.2453} & \textbf{33.3680} & \textbf{0.9374} & \textbf{0.8441} & 79.9 & \textbf{32.5976} & \textbf{4.8720} & 79.3\\
OursOF & 0.4683 & 26.1421 & 0.6364 & 0.6741 & \textbf{86.7} & 28.3269 & 4.2194 & \textbf{83.9}\\
\hline
\end{tabular}
\end{center}
\end{table}
\begin{figure*}[h]
\centering
\includegraphics[height=8cm, width=18cm]{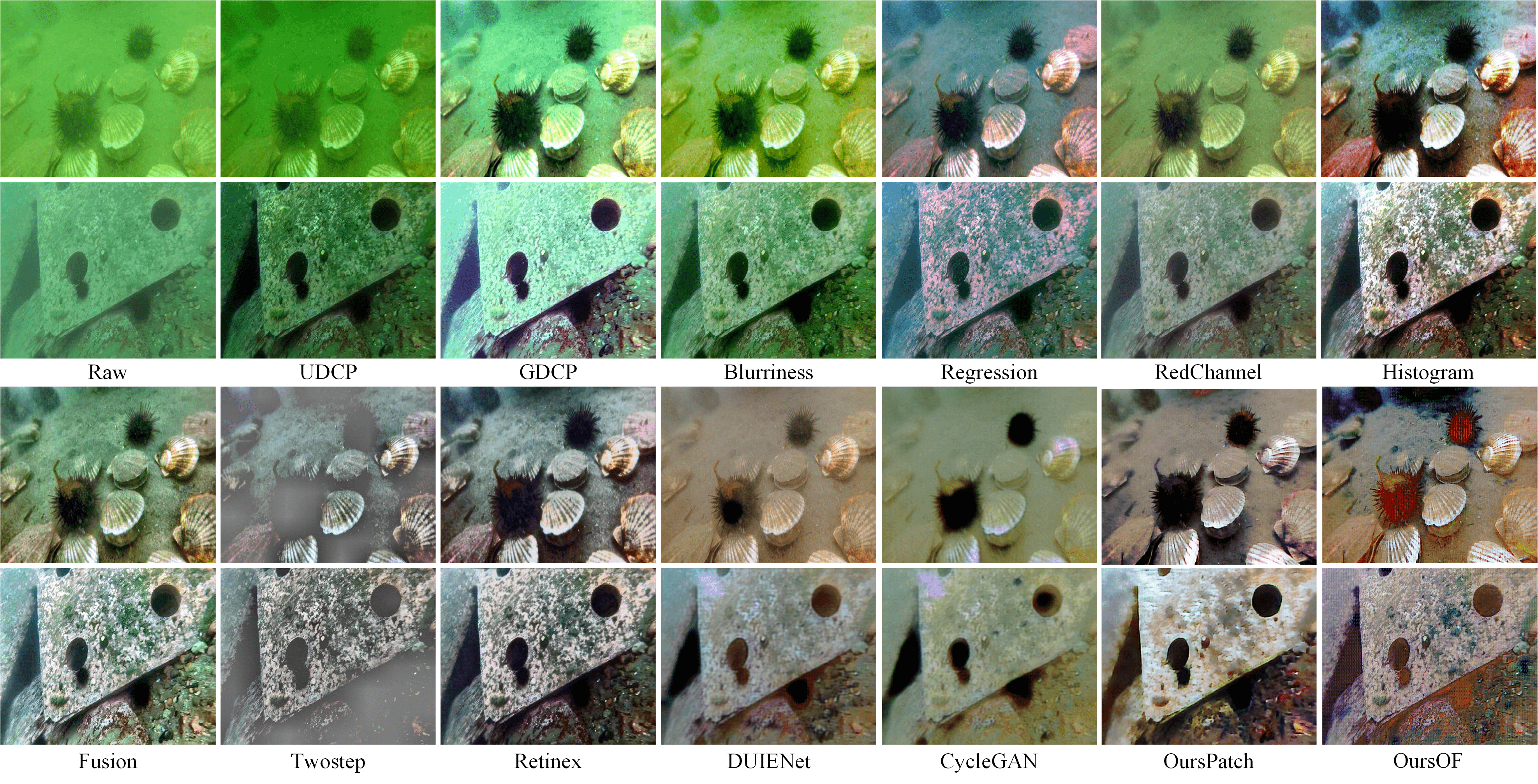}
\caption{Qualitative comparison of different UIE algorithms on the ChinaMM dataset. From left to right are raw underwater images, results of UDCP \cite{b16}, GDCP \cite{b17}, Blurriness \cite{b2}, Regression \cite{b20}, RedChannel \cite{b18}, Histogram \cite{b21}, Fusion \cite{b13}, Two-step \cite{b14}, Retinex \cite{b15}, DUIENet \cite{b22}, CycleGAN \cite{b12}, OursPatch and OursOF. The top raw is greenish and the bottom one is bluish.}
\label{fig:sot_enhance_ChinaMM}
\end{figure*}
\begin{figure}[h]
\centering
\includegraphics[height=4.5cm, width=9cm]{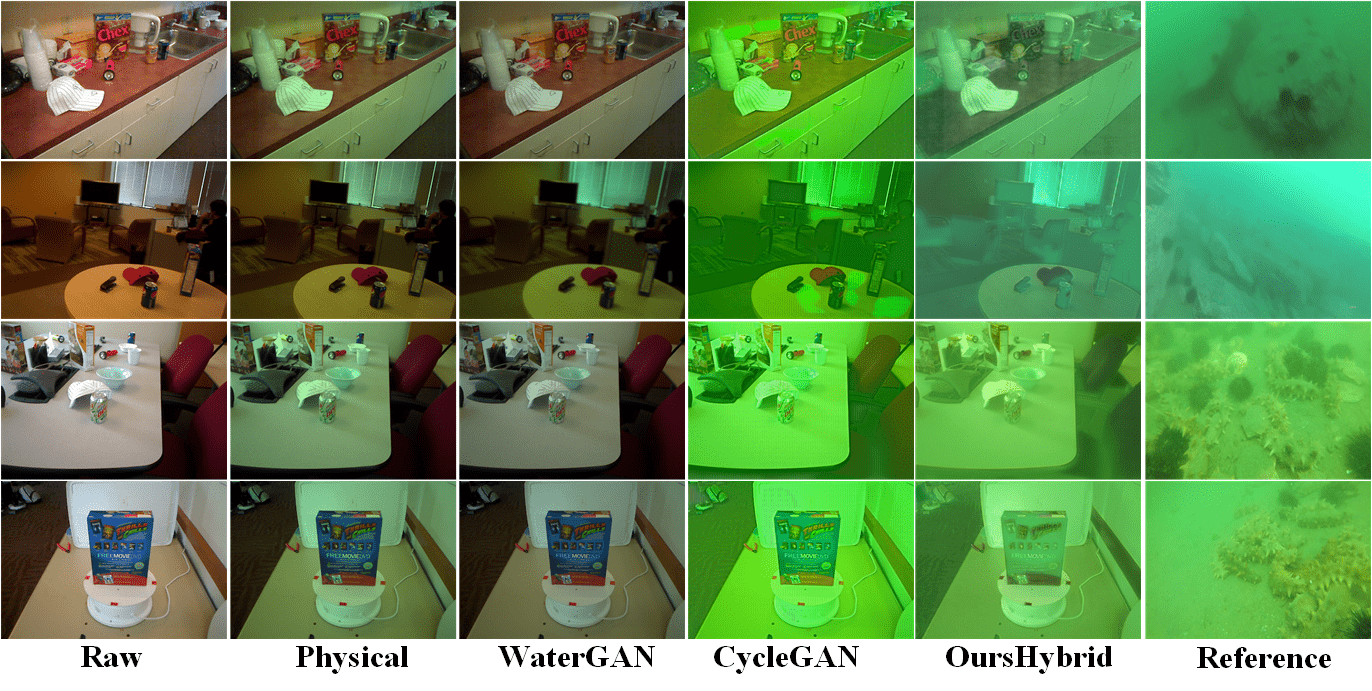}
\caption{Qualitative comparison of different UIS algorithms on the MultiView dataset. From left to right are raw in-air images of MultiView, results of Physical \cite{b1}, WaterGAN \cite{b10}, CycleGAN \cite{b12}, OursHybrid, and real underwater images from ChinaMM as the references. Best viewed in the digital form.}
\label{fig:sot_syn_Multiview}
\end{figure}
Table~\ref{table:quantiMVU_det} reports the quantitative results of the enhancement models with different detection perceptor settings on MultiViewUnderwater and ChinaMM. On MultiviewUnderwater, OursPatch achieves the best full-reference scores, and the corresponding deep detector obtains almost the same mAP as the on trained on the ground-truth in-air images of MulitiView. This quantitative performance attributes to its enhanced results similar to the ground-truth in-air images. The deep detector trained on the enhanced results of OursOF achieves the best mAP. We have similar experimental results on ChinaMM, where OursPatch achieves the best UCIQE score (32.5976) and UIQM score (4.8720), whilst OursOF achieves the best mAP (83.9). We believe the best detection accuracy is due to the reduction of the disturbing background. To verify this assumption, we use the detection tool of \cite{b45} to analyse the false positives of the two detectors trained on the results of OursPatch and OursOF. Fig.~\ref{fig:error} and Supplementary F show the distribution of the top-ranked false positive measures for each category on the testing sets of ChinaMM and MultiviewUnderwater. The former detector cannot well distinguish the objects with complex background while the latter one largely reduces the background errors. The qualitative and quantitative comparisons of the enhancement models with different detection perceptor settings on the paired OUC dataset can be found in Supplementary E.

\subsection{Comparison with state-of-the-art methods}
In this subsection, we first compare our hybrid synthesis model with three state-of-the-art UIS algorithms. Then, we compare our two detection based perceptual enhancement models with eleven state-of-the-art UIE algorithms.\\
\textbf{Comparison with state-of-the-art UIS methods.} We compared our hybrid synthesis model (denoted as OursHybrid) with three state-of-the-art UIS algorithms, including physical model-based UIS method (denoted as Physical) \cite{b1}, CycleGAN \cite{b12} and WaterGAN \cite{b10}. Physical \cite{b1} applied the physical underwater image formation model and 10 groups of pre-defined parameters to synthesise 10 Jerlov type underwater images from RGB-D in-air images, and the synthetic dataset contains 10 types of underwater images with various color distortions and haze-effects. We choose the images most similar to the ChinaMM-style underwater images in the qualitative comparison. WaterGAN and CycleGAN are deep learning based UIS methods, we tune their parameter settings to generate satisfactory results.

\begin{table*}
\begin{center}
\caption{Non-Reference image quality and detection accuracy evaluations on the ChinaMM dataset.}
\label{table:quantiChinaMM_perceptor}
\resizebox{\textwidth}{12mm}{
\begin{tabular}{ccccccccccccccc}
\hline\noalign{\smallskip}
Metrics & UDCP & GDCP & Blurriness & Regression & RedChannel &  Histogram & Fusion & Twostep & Retinex & DUIENet & CycleGAN & OursPatch & OursOF\\
\noalign{\smallskip}
\hline
\noalign{\smallskip}
UCIQE & 28.6184 & \textbf{33.6328} & 30.5865 & 29.3877 & 30.8712 & 33.3443 & 31.7698 & 15.1238 & 28.447 & 31.5588 & 30.7922 & 32.5976 & 32.0967\\
UIQM & 3.0184 & 2.6468 & 3.6984 & 3.7616 & 3.3028 & 4.6728 & 4.0696 & 2.6728 & 4.7306 & 2.7021 & 3.7036 & \textbf{4.8720} & 4.4621\\
UICM & -56.7266 & -53.7373 & -58.1656 & -21.9885 & -31.2248 & 0.4558 & -22.2500 & -4.3928 & -0.4811 & -39.4952 & 6.9885 & \textbf{13.9967} & 12.4050\\
UISM & 6.7033 & \textbf{6.7830} & 6.7155 & 6.7060 & 6.6955 & 6.7210 & 6.6525 & 5.7978 & 6.6751 & 6.6242 & 6.5753 & 6.7326 & 6.5311\\
UICONM & 0.7380& 0.6039 & \textbf{0.9385} & 0.6716 & 0.6170 & 0.7482 & 0.7643 & 0.3033 & 0.7755 & 0.5201 & 0.4376 & 0.6962 & 0.6107\\
mAP & 71.6 & 72.7 & 77.3 & 71.6 & 73.7 & 76.8 & 75.5 & 58.6 & 78.8 & 71.6 & 67.8 & 79.3 & \textbf{83.9}\\
\hline
\end{tabular}}
\end{center}
\end{table*}
Fig.~\ref{fig:sot_syn_Multiview} shows the qualitative comparison of different UIS algorithms on MultiView. It is evident that the resultant images of our proposed hybrid synthesis model are very similar to the reference images of ChinaMM in terms of diversity, color casts and haze-effects. In contrast, the results of WaterGAN suffer from insufficient haze-effects due to the lack of light scattering prior, even though they apply a shallow convolutional network to simulating the light scattering process. However, in practice, without referring to the optical property of light scattering, the ability of CNN to simulate haze-effects is limited. In addition, the results of WaterGAN suffer from unrealistic color distortion even though  it has used the light absorption prior. This is mainly because the weekly supervised WaterGAN is trained with only an adversarial loss on the unpaired images, which cannot provide sufficient supervision information to simulate color distortion. Apart from the adversarial loss, CycleGAN adds a pixel-wise cycle-consistency loss as an additional constraint to supervising the training, and the cycle-consistency loss to minimise the discrepancy between the generated and the ground-truth images in the pixel level that leads to more realistic color distortion. Nevertheless, evident artifacts and no haze-effects appear in the results of CycleGAN because the CNN structure has the limited capability to simulate haze-effects and remove artifacts. It is also worth noting that WaterGAN and CycleGAN generate underwater images in a monotonous style due to the model collapse problem, restricting their generalisation capability to yield diverse real-world underwater images. By integrating the physical prior into CNN, OursHybrid can generate images with diverse properties. Physical is able to generate artifact-free results but cannot simulate realistic color distortion as it uses the fixed parameters defined by the Jerlov water type. The qualitative and quantitative comparison of different UIS methods on OUC can be found in Supplementary G.\\
\textbf{Comparison with state-of-the-art UIE methods.} We compare our two detection perceptual enhancement models with eleven state-of-the-art UIE algorithms, including six physical model-based methods (i.e., UDCP \cite{b16}, GDCP \cite{b17}, Blurriness \cite{b2}, Regression \cite{b20}, RedChannel \cite{b18} and Histogram \cite{b21}), three model-free methods (i.e., Fusion \cite{b13}, Two-step \cite{b14}, and Retinex \cite{b15}), and two deep learning-based methods (i.e., DUIENet \cite{b22} and CycleGAN \cite{b12}) on three datasets (i.e., synthetic MultiviewUnderwater, real-world ChinaMM and OUC).

Fig.~\ref{fig:sot_enhance_ChinaMM} and Supplementary H Fig.~\ref{fig:sot_enhance_Multiview} show the qualitative comparison of different UIE algorithms on ChinaMM and MultiviewUnderwater, respectively. For the two datasets, most of the physical model-based UIE algorithms cannot deal with severe color distortions. Among them, Histogram performs relative better for color distortions which benefits from the histogram prior it uses. But it generates over-saturation and excessive contrast in some image regions. UDCP and Blurriness even aggravate the color distortion due to the limitations of the priors used in these two methods. Regression tends to introduce bluish color casts on account of its inaccurate color correction algorithm, and GDCP over enhances the brightness that results in the loss of image details. RedChannel improves little on the color distortion. Among the model-free algorithms, Retinex can effectively remove color distortion and produce more natural scenes while Fusion improved little the color distortion. Twostep over-enhances the contrast of the underwater images and generate unnatural results. The deep learning based methods DUIENet and CycleGAN can deal with the color casts, however, both of them still leave evident haze on the resultant images. In terms of haze removal, most of the physical model-based methods are able to remove the haze-effects to some extend, benefiting from the use of light scattering priors. Among the model-free methods, Retinex and Fusion effectively remove the haze-effects on the underwater images while Twostep does little about the haze removal. Among all the UIE methods, OursPatch achieves the best qualitative performance in terms of color tones, visibility, saturation and contrast. The qualitative comparison of different UIE algorithms on OUC can be found in Supplementary H Fig.~\ref{fig:sot_enhance_OUC_haze}.

\begin{figure}[h]
\centering
\includegraphics[height=2.8cm, width=9cm]{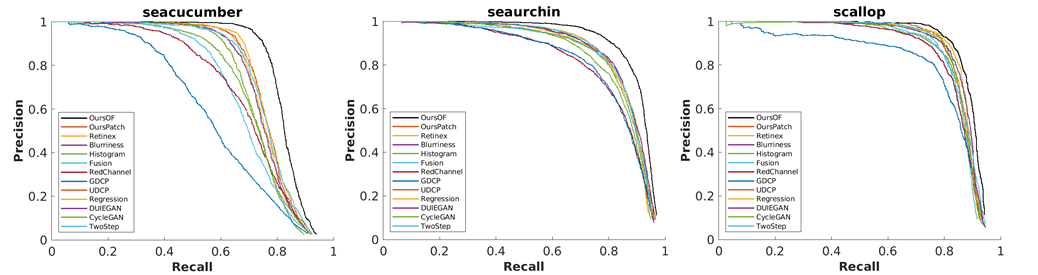}
\caption{Precision/Recall curves of deep detectors trained on the results of different UIE methods on ChinaMM.}
\label{fig:rocChinaMM}
\end{figure}
In addition to qualitative evaluations, we also report the quantitative results of different UIE algorithms. The best values are marked in bold. The quantitative results on the testing set of MultiviewUnderwater and OUC are presented in Supplementary I, from which we can see that OursPatch achieves the lowest MSE score and the highest scores of PSNR, SSIM and PCQI. A higher PSNR score and a lower MSE score denote that the result is closer to the reference image in terms of image content, while a higher SSIM score means the result is more similar to the reference image in terms of image structure and texture. Table~\ref{table:quantiChinaMM_perceptor} gives the quantitative scores of non-reference metrics, i.e., UIQM, UCIQE, and three components of UIQM (UICM, UISM, and UIConM). The following algorithms perform over others in terms of one single metric: GDCP achieves the best UCIQE score, and Blurriness achieves the best UICONM score. However, the results of both suffer from serious color casts as shown in Figs.~\ref{fig:sot_enhance_Multiview} and \ref{fig:sot_enhance_ChinaMM}. There exists discrepancies between the qualitative images and the quantitative scores in some cases which has also been verified in \cite{b1}. For mAP, the Precision/Recall curves in Fig.~\ref{fig:rocChinaMM} and Supplementary H show that OursOF (black curve) performs the best across all categories on MultiviewUnderwater and ChainMM, which indicates the interaction between enhancement model and the detection perceptor brings significant performance improvement for the following deep detector.

\subsection{The influences of UIE algorithms on the detection task.} Previous works seem to suggest that UIE algorithms will bring improvements of image quality, which further boosts the performance of the following high-level detection tasks. We conduct the following analysis to investigate whether or not UIE brings improvement of detection accuracy.

\begin{figure}[h]
\centering
\includegraphics[height=3.8cm, width=9cm]{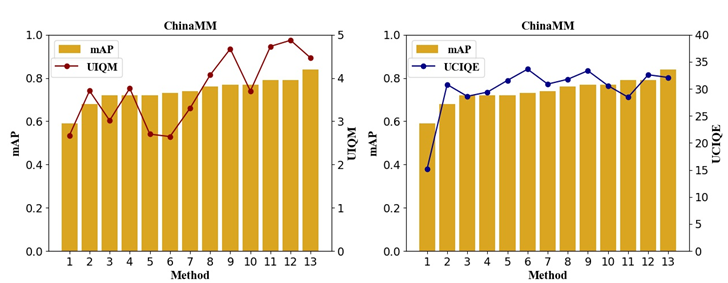}
\caption{Image quality evaluation metrics and mAP on ChinaMM. The histogram represents the mAP and the polyline represents different image quality evaluation metrics. Numbers 1 to 13 refer to thirteen UIE algorithms ordered according to increasing mAP values.}
\label{fig:metricsrelationChinaMM}
\end{figure}
We compare the quantitative scores of the raw underwater images and the enhanced underwater images by different UIE algorithms on the three underwater datasets. It is observed that not all the UIE algorithms increase the quantitative scores of raw underwater images. Deep learning based methods can stably improve image quality on both full-reference and non-reference metrics, while other methods only work for some special image quality evaluation metrics. 

Fig.~\ref{fig:metricsrelationChinaMM} and Supplementary J illustrate mAP and image quality evaluation metrics on ChinaMM and MultiviewUnderwater datasets, from which we investigate how the detection accuracy is related to different image quality evaluation metrics. There is no strong correlation  between the mAP and the image quality evaluation metrics on the two datasets. For MultiviewUnderwater, Regression receives the best MSE and PCQI scores among the six physical model-based methods, however, its detection accuracy is the worst among these methods. For ChinaMM, CycleGAN can greatly improve the UCIQE and UIQM scores, but its mAP (67.8\%) is even lower than that of the raw underwater images (68.4\%). On the OUC dataset, both GDCP and UDCP decrease the MSE and PSNR scores of the raw underwater images, but their mAP are even higher than these of the high quality reference images (86.6\% mAP). Therefore, discrepancies exist between the image quality evaluation metrics and the detection accuracy. mAP may be biased to some image quality metrics such as high UICONM and UICM. For example, Blurriness, Histogram, Fusion and Retinex rank top four UICONM scores, and achieve the top four detection accuracy among the non-deep learning based methods. OursPatch and OursOF receive significantly higher UICM  scores and rank top two in terms of mAP. The detection task tends to favour the results with high contrasts (high UICONM) between the objects and the background, or that with over-enhanced objects. One possible explanation lies in that high contrast suppresses the complicated background while bright color protrudes the objects. For illustration, we show the object detection results of two underwater images from MultiviewUnderwater and ChinaMM in Supplementary K.

We believe there is a gap between the image quality evaluation metrics for the low-level enhancement task and the accuracy metric for the high-level detection task. Underwater image enhancement task is usually evaluated using the image quality evaluation metrics. However, the objective of the low-level enhancement task typically differs from that of the high-level object detection task so that enhancement algorithm can hardly recover features favoured by the high-level detection task. The interaction between the enhancement and detection tasks are important for the development of the two tasks in the future.

\section{Conclusion}
In this paper, we have proposed two detection-perceptual enhancement models, i.e., patch detection-based perceptual enhancement and object-focused detection based perceptual enhancement models. With the help of two detection perceptors, our patch detection-based perceptual enhancement model can generate high quality in-air images with patch-level details, and our object-focused detection-based perceptual enhancement model can generate images which improves the detection accuracy of the following deep detectors. Moreover, to advance the generalisation ability of deep learning based UIE algorithms, we have proposed a hybrid underwater image synthesis model to synthesise more realistic training images, and the enhancement model trained on them can learn more robust translation between the underwater and high quality in-air images, and generalise well on the real-world underwater scenes. The synthesis model, enhancement model and detection perceptor are trained in a unified HybridDetectionGAN framework, which is superior to several state-of-the-art techniques.

\clearpage
\onecolumn
\setcounter{table}{0}
\setcounter{figure}{0}
\setcounter{equation}{0}
\setcounter{page}{1}
\renewcommand\thefigure{S\arabic{figure}} 
\renewcommand\thetable{S\arabic{table}} 
\section*{Supplementary A}
\begin{figure*}[htb]
\centering
\includegraphics[height=7cm, width=18cm]{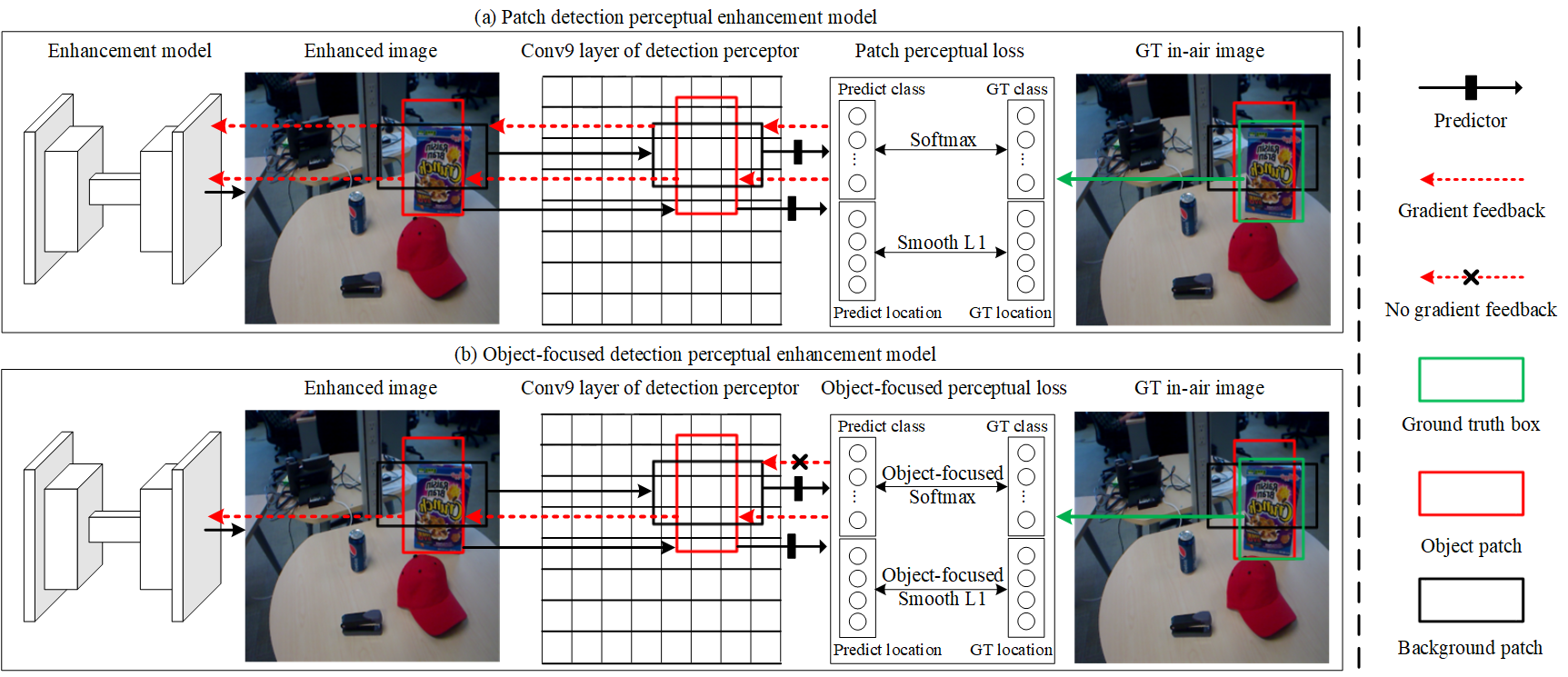}
\caption{The framework of (a) patch detection perceptual enhancement model and (b) object-focused detection perceptual enhancement model. The perceptual loss computes the discrepancy between patches on the enhanced image and that on the ground truth in-air image and feeds back this discrepancy to update the enhancement model in the form of gradients. The patch detection perceptual enhancement model feeds back informations of the background and object patches, while the object-focused detection perceptual enhancement model only feeds back the object information.}
\label{fig:dpenhancement}
\end{figure*}
\section*{Supplementary B}
The physical parameters $\alpha$, $\beta$ and $B$ are achieved via the gradient descent optimisation algorithm, which iteratively updates these parameters by minimising $L_{a2w}$. The optimisation algorithm iteratively updates the parameter $\beta^{\lambda}$ by
\begin{equation}
\begin{split}
\beta^{\lambda}=\beta^{\lambda}-\eta\frac{\partial L_{a2w}}{\partial \beta^{\lambda}}
\end{split}
\label{eq:19}
\end{equation}
where $\eta$ is the learning rate. In order to update $\beta^{\lambda}$, we need to compute $\frac{\partial L_{a2w}}{\partial \beta}$, which indicates the gradient of $L_{a2w}$ with respect to $\beta^{\lambda}$. It is derived from the following chain rule: 
\begin{equation}
\begin{aligned}
\frac{\partial L_{a2w}}{\partial \beta^{\lambda}}&= \frac{\partial L_{a2w}}{\partial L_{adv\_w}}\frac{\partial L_{adv\_w}}{\partial \beta^{\lambda}}+  \frac{\partial L_{a2w}}{\partial L_{cyc\_w}}\frac{\partial L_{cyc\_w}}{\partial \beta^{\lambda}}\\
&=w_1\frac{\partial L_{adv\_w}}{\partial \beta^{\lambda}}+ w_2\frac{\partial L_{cyc\_w}}{\partial \beta^{\lambda}}
\end{aligned}
\label{eq:20}
\end{equation}
We denote $I_{add}^{\lambda}$ as the output of the physical branch, $I_{add}^{\lambda}=I_{ab}^{\lambda}+I_{sc}^{\lambda}$. $I_{ab}^{\lambda}$, $I_{con}$ and $I_{sw}$ are formulated in Eqs.~(\ref{eq:newphysical})-(\ref{eq:absorption}). We derive $\frac{\partial L_{adv\_w}}{\partial \beta^{\lambda}}$ and $\frac{\partial L_{cyc\_w}}{\partial \beta^{\lambda}}$ using the chain rule as follows:
\begin{equation}
\begin{aligned}
\frac{\partial L_{adv\_w}}{\partial \beta^{\lambda}}&=\frac{\partial L_{adv\_w}}{\partial I_{sw}} \frac{\partial I_{sw}}{\partial I_{con}} \frac{\partial I_{con}}{\partial I_{add}^{\lambda}} \frac{\partial I_{add}^{\lambda}}{\partial I_{ab}^{\lambda}} \frac{\partial I_{ab}^{\lambda}}{\partial \beta^{\lambda}}\\
&=-I_{ab}^{\lambda}\otimes e^{-I_d \beta^{\lambda}} I_{d}\frac{\partial L_{adv\_w}}{\partial I_{sw}} \frac{\partial I_{sw}}{\partial I_{con}} \frac{\partial I_{con}}{\partial I_{add}^{\lambda}}
\end{aligned}
\label{eq:21}
\end{equation}
\begin{equation}
\begin{aligned}
\frac{\partial L_{cyc\_w}}{\partial \beta^{\lambda}}&=\frac{\partial L_{cyc\_w}}{\partial I_{sw}} \frac{\partial I_{sw}}{\partial I_{con}} \frac{\partial I_{con}}{\partial I_{add}^{\lambda}} \frac{\partial I_{add}^{\lambda}}{\partial I_{ab}^{\lambda}} \frac{\partial I_{ab}^{\lambda}}{\partial \beta^{\lambda}}\\
&=-I_{ab}^{\lambda}\otimes e^{-I_d \beta^{\lambda}} I_{d}\frac{\partial L_{cyc\_w}}{\partial I_{sw}} \frac{\partial I_{sw}}{\partial I_{con}} \frac{\partial I_{con}}{\partial I_{add}^{\lambda}}
\end{aligned}
\label{eq:22}
\end{equation}
Combining Eqs.~(\ref{eq:20})-(\ref{eq:22}) with Eq.~(\ref{eq:19}),  we have
\begin{equation}
\begin{aligned}
\beta^{\lambda}=&\beta^{\lambda}+\eta(w_1\frac{\partial L_{adv\_w}}{\partial I_{sw}}+w_2\frac{\partial L_{cyc\_w}}{\partial I_{sw}})\\
&(I_{ab}^{\lambda}\otimes e^{-I_d \beta^{\lambda}}I_d \frac{\partial I_{sw}}{\partial I_{con}} \frac{\partial I_{con}}{\partial I_{add}^{\lambda}})
\end{aligned}
\end{equation}
Similarly, we update $\alpha$ and $B$ through the chain rule:
\begin{equation}
\begin{aligned}
\alpha^{\lambda}=&\alpha^{\lambda}-\eta(w_1\frac{\partial L_{adv\_w}}{\partial I_{sw}}+w_2\frac{\partial L_{cyc\_w}}{\partial I_{sw}})\\
&(B^{\lambda} e^{-I_d \alpha^{\lambda}}I_d \frac{\partial I_{sw}}{\partial I_{con}} \frac{\partial I_{con}}{\partial I_{add}^{\lambda}})
\end{aligned}
\end{equation}
\begin{equation}
\begin{aligned}
B^{\lambda}=&B^{\lambda}-\eta(w_1\frac{\partial L_{adv\_w}}{\partial I_{sw}}+w_2\frac{\partial L_{cyc\_w}}{\partial I_{sw}})\\
&((1-e^{-I_d \alpha^{\lambda}}) \frac{\partial I_{sw}}{\partial I_{con}} \frac{\partial I_{con}}{\partial I_{add}^{\lambda}})
\end{aligned}
\end{equation}
\section*{Supplementary C}
\begin{figure*}[h]
\centering
\includegraphics[height=6cm, width=18cm]{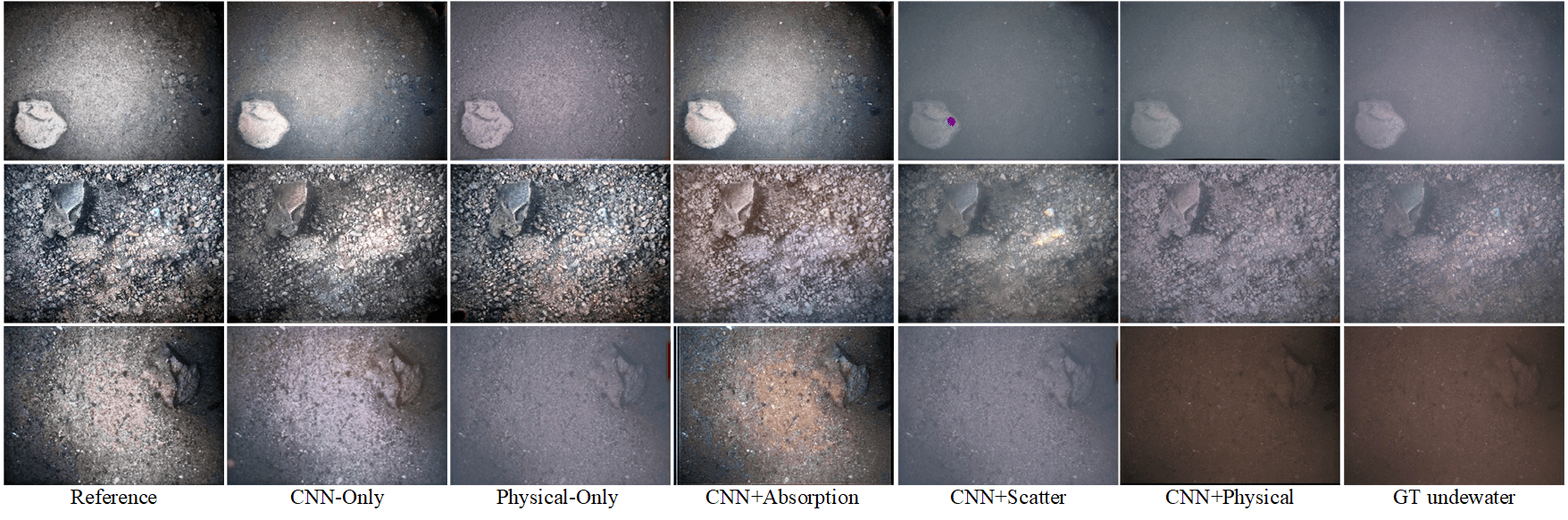}
\caption{Qualitative comparison of the synthesis models with different component settings on OUC. From left to right are high quality reference images, the results of the synthesis models with only CNN branch, only physical branch, CNN and absorption branches, and CNN and scattering branches, complete hybrid synthesis model, and the ground-truth underwater images.}
\label{fig:syn_OUC}
\end{figure*}

\section*{Supplementary D}
\begin{figure*}[h]
\centering
\includegraphics[height=8cm, width=18cm]{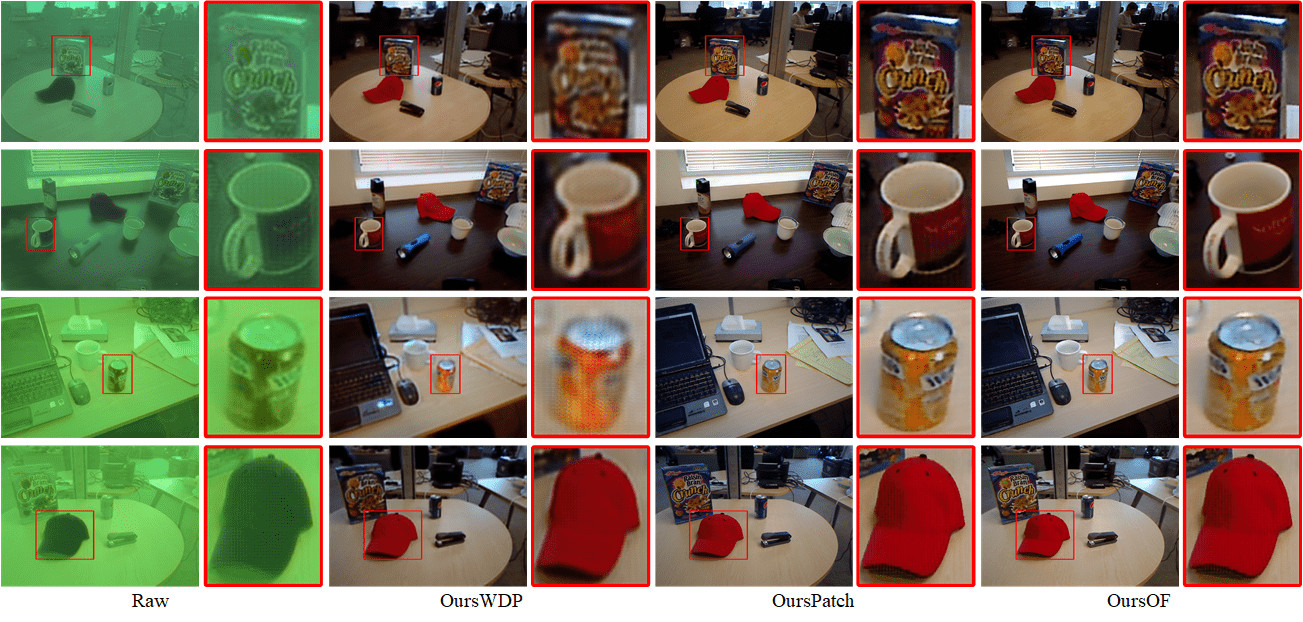}
\caption{Qualitative comparison of the enhancement models with different detection perceptor settings on MultiviewUnderwater. From left to right are raw underwater images, results of OursWDP, OursPatch, and OursOF, respectively.}
\label{fig:ablation_enhance_Multiview_perceptor}
\end{figure*}

\section*{Supplementary E}
\begin{figure*}[h]
\centering
\includegraphics[height=6cm, width=18cm]{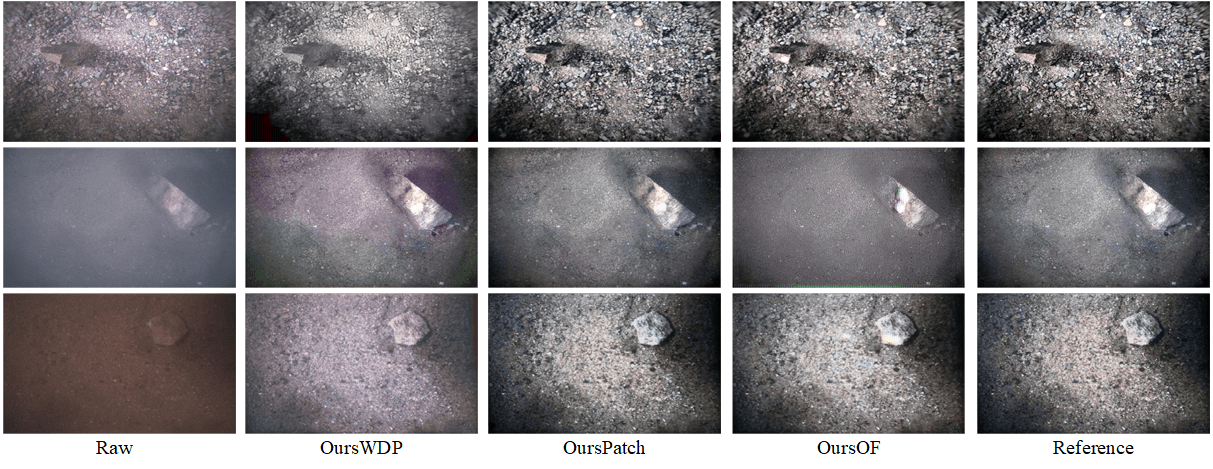}
\caption{Qualitative comparison of the synthesis models with different perceptor settings on the OUC dataset.}
\label{fig:ablation_enhance_OUC}
\end{figure*}

From Fig.~\ref{fig:ablation_enhance_OUC}, we observe that resultant images of OursWDP still suffer artifacts and haze-effects, and the patch detection perceptor helps OursPatch to remove these negative effects and recover more image details. OursOF somehow over-enhanced the objects but under-enhanced the background. In Table~\ref{table:quantiSyn_OUC_det}, OursPatch outperforms the other two models in terms of the four full-reference metrics. OursOF greatly improves the detection accuracy of the late detector with favourable synthesis images.
\begin{table*}[h]
\begin{center}
\caption{Quantitative comparison of the enhancement models with different detection perceptor settings on the OUC dataset.}
\label{table:quantiSyn_OUC_det}
\begin{tabular}{llllll}
\hline
Models & MSE & PSNR & SSIM & PCQI & mAP\\
\hline
\noalign{\smallskip}
OursWDP & 0.1168   & 28.3034 & 0.85821 & 0.7147 & 83.5\\
OursPatch & \textbf{0.0224}  & \textbf{35.4039} & \textbf{0.9724} & \textbf{0.9389} & 86.5\\
OursOF & 0.0616 & 30.8662 & 0.9351 & 0.9030 & \textbf{90.1}\\
\hline
\end{tabular}
\end{center}
\end{table*}

\section*{Supplementary F}
\begin{figure*}[h]
\centering
\includegraphics[height=8cm, width=18cm]{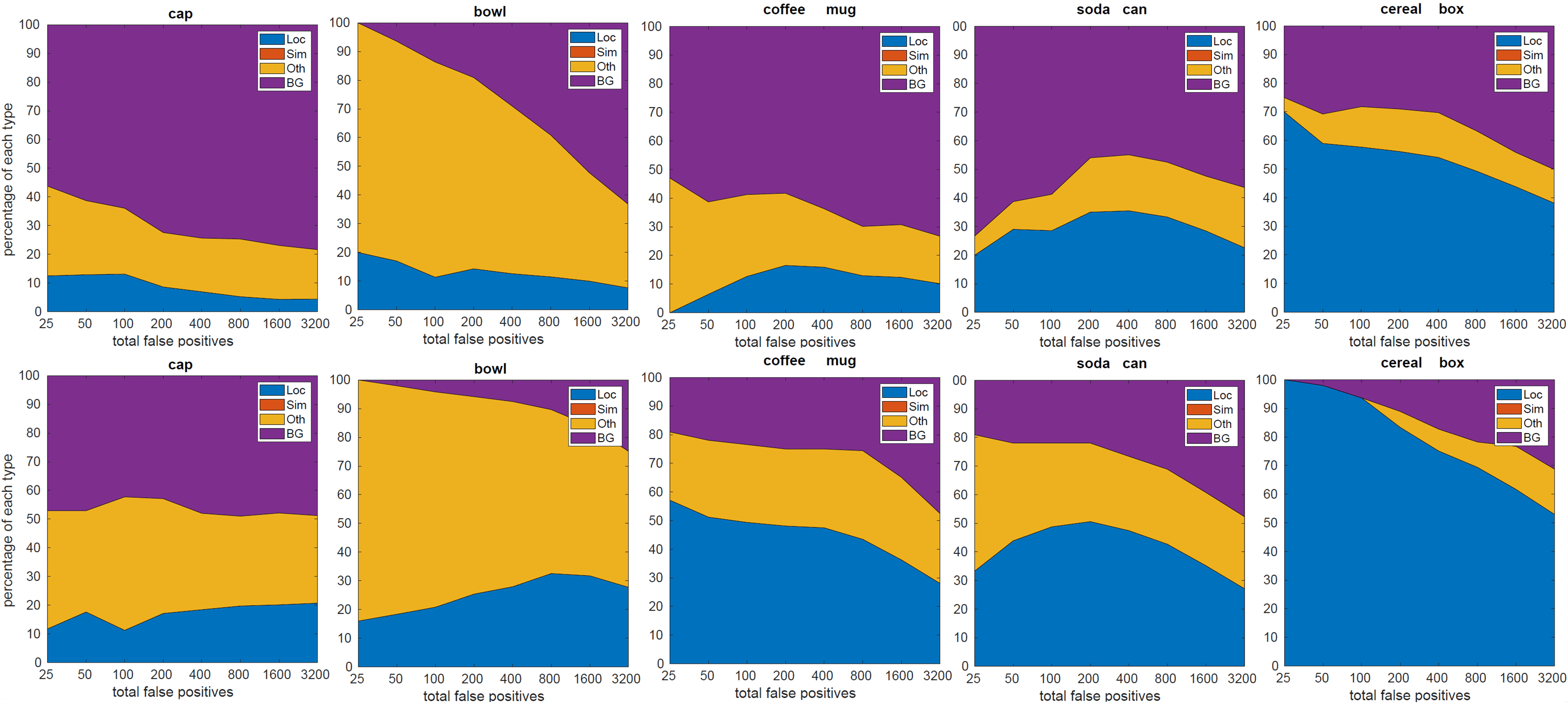}
\caption{The distribution of the top-ranked false positive measures for images of MultiviewUnderwater. The false positive measures include localisation errors (Loc), confusion with similar categories (Sim), with others (Oth), or with background (BG). The top row shows the results of OursPatch and the bottom row shows the results of OursOF.}
\label{fig:errorMultiviewUnderwater}
\end{figure*}

\section*{Supplementary G}
\begin{figure*}[h]
\centering
\includegraphics[height=8cm, width=18cm]{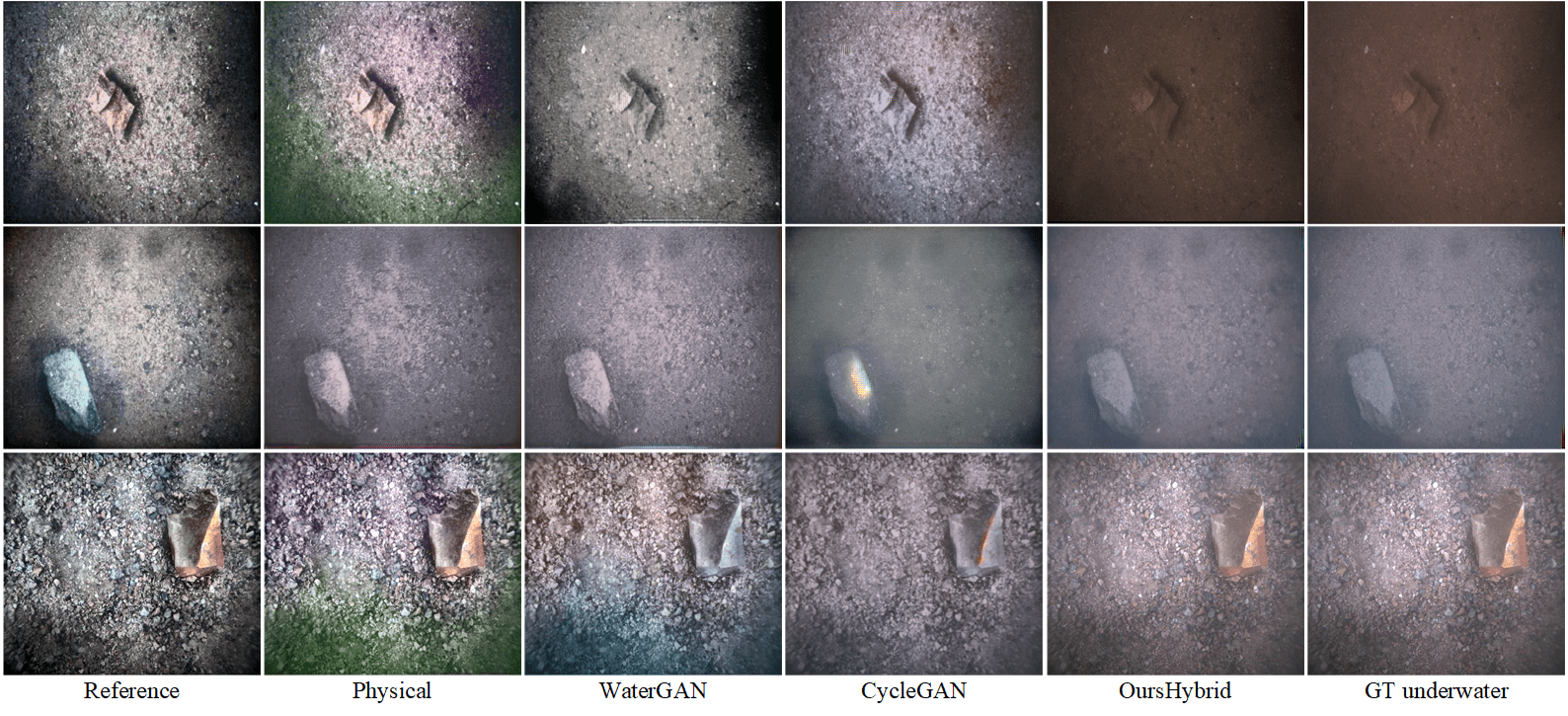}
\caption{Qualitative comparison of different UIS algorithms on the OUC dataset. From left to right are high-quality reference images, results of Physical \cite{b1}, WaterGAN \cite{b10}, CycleGAN \cite{b12}, OursHybrid, and ground-truth underwater images.}
\label{fig:sot_syn_OUC}
\end{figure*}

From Fig.~\ref{fig:sot_syn_OUC} we observe that Physical cannot generate suitable color distortions while WaterGAN and CycleGAN generate images with monotonous color tones. OursHybrid is able to generate underwater images with diverse color casts and evident haze-effects. Table~\ref{table:quantiOUC_sot_syn} reports the average MSE, PSNR, SSIM and PCQI scores of four UIS methods. OursHybrid outperforms the other three methods in terms of four full-reference metrics.

\begin{table}[h]
\begin{center}
\caption{Full-Reference image quality evaluations of different UIS algorithms on the OUC dataset.}
\label{table:quantiOUC_sot_syn}
\begin{tabular}{lllll}
\hline\noalign{\smallskip}
Methods & MSE & PSNR & SSIM & PCQI\\
\noalign{\smallskip}
\hline
\noalign{\smallskip}
Physical & 0.4450 & 21.7190 & 0.6849 & 0.4782\\
WaterGAN & 0.3112 & 23.2620 & 0.7378 & 0.5916\\
CycleGAN & 0.3007 & 23.3956 & 0.7549 & 0.6384\\
OursHybrid & \textbf{0.0978} & \textbf{28.2895} & \textbf{0.9131} & \textbf{0.9518}\\
\hline
\end{tabular}
\end{center}
\end{table}

\section*{Supplementary H}
\begin{figure*}[h]
\centering
\includegraphics[height=3.5cm, width=18cm]{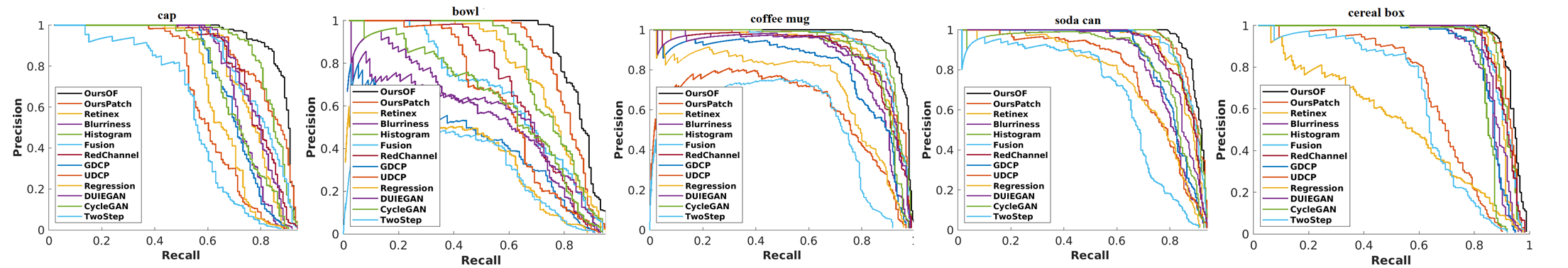}
\caption{Precision/Recall curves of different methods on ChinaMM (seacucumber, seaurchin, and scallop) and MultiviewUnderwater (ceal box, soda can, coffee mug, cap and bowl).}
\label{fig:rocpaired}
\end{figure*}

\begin{figure*}[h]
\centering
\includegraphics[height=9cm, width=18cm]{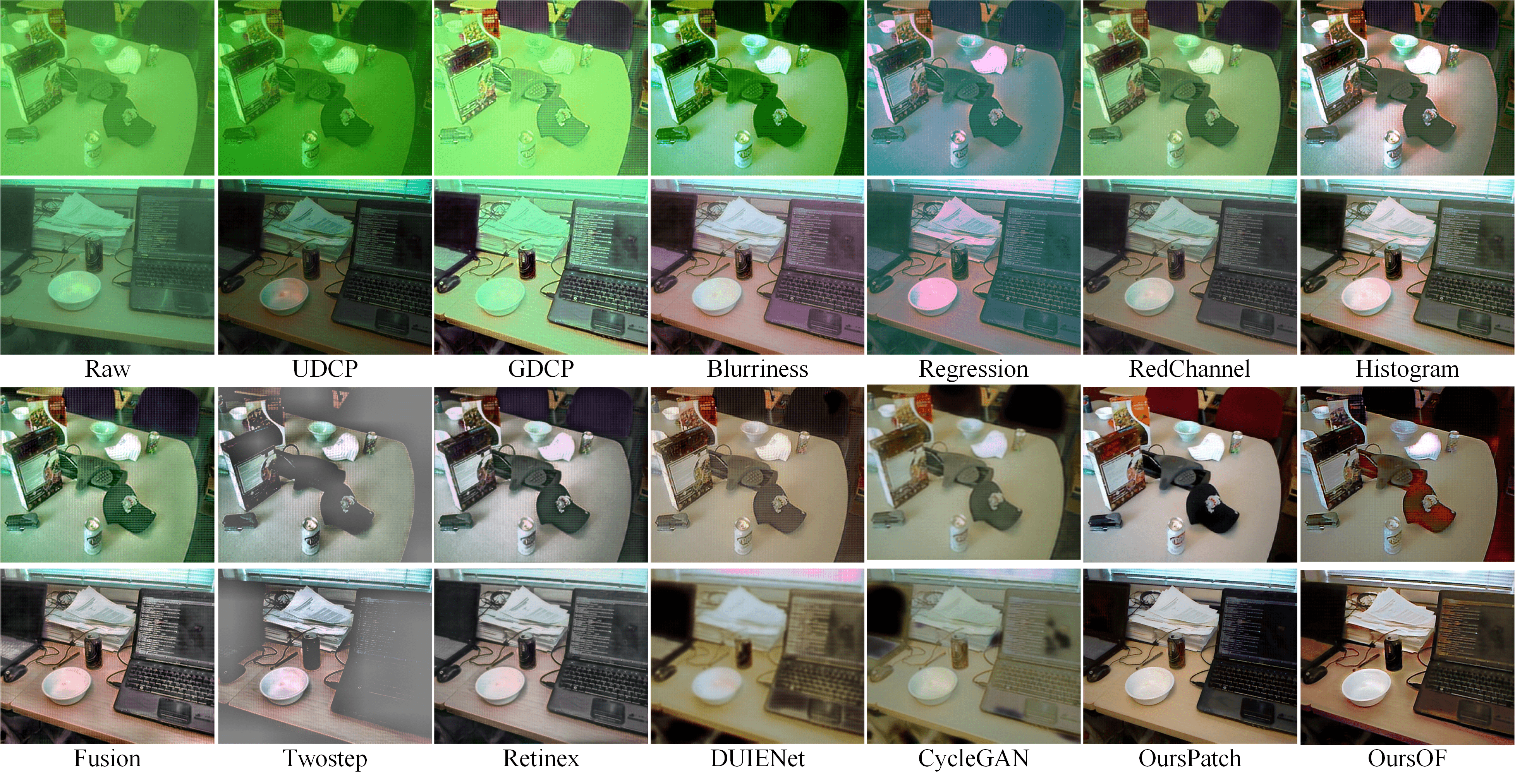}
\caption{Qualitative comparison of different UIE algorithms on the MultiviewUnderwater dataset. From left to right are raw images, results of UDCP \cite{b16}, GDCP \cite{b17}, Blurriness \cite{b2}, Regression \cite{b20}, RedChannel \cite{b18}, Histogram \cite{b21}, Fusion \cite{b13}, Two-step \cite{b14}, Retinex \cite{b15}, DUIENet \cite{b22}, CycleGAN \cite{b12}, OursPatch and OursOF. The top raw is greenish and the bottom one is bluish.}
\label{fig:sot_enhance_Multiview}
\end{figure*}

\begin{figure*}[h]
\centering
\includegraphics[height=9.7cm, width=18cm]{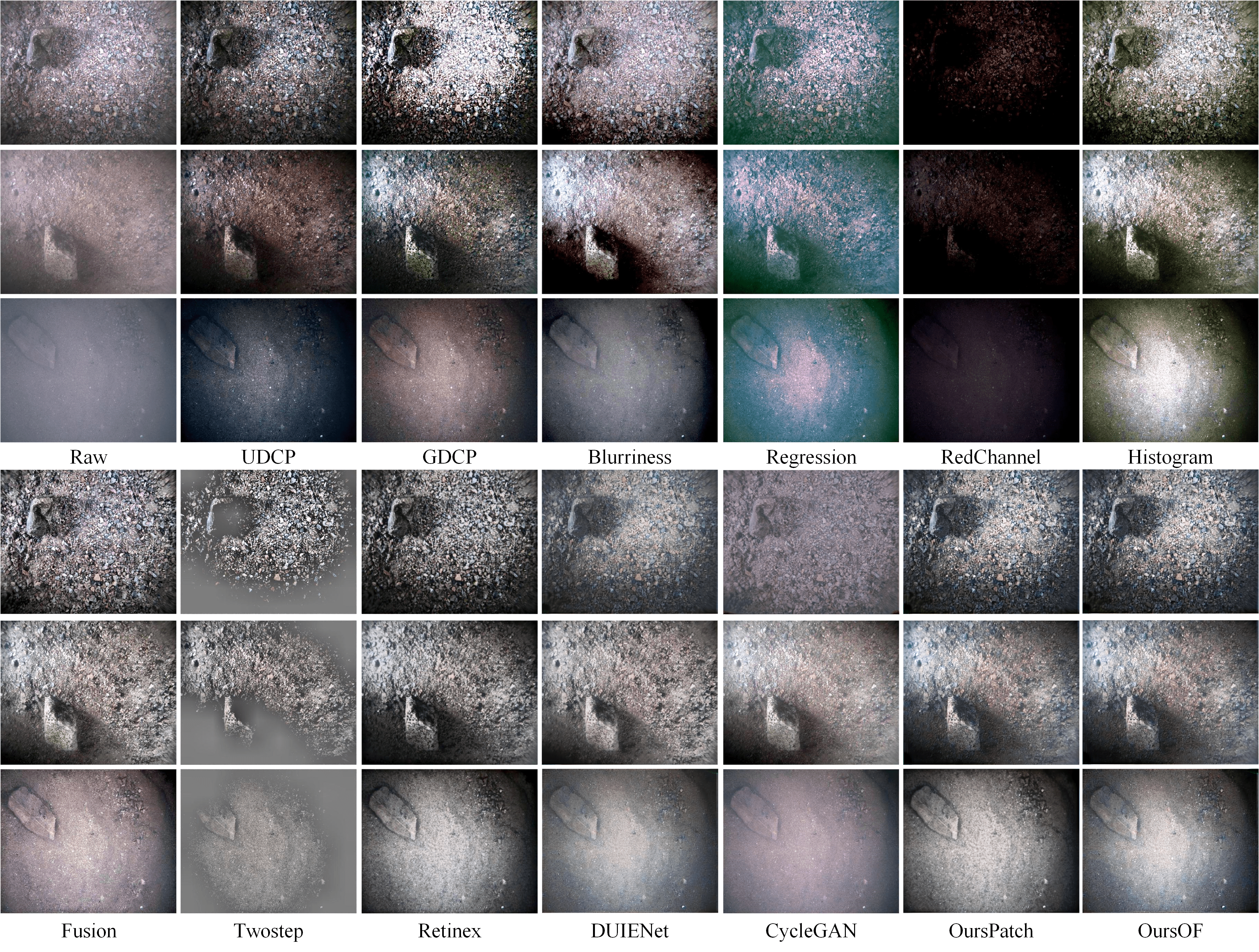}
\caption{Qualitative comparison of different UIE methods on the OUC images under different haze levels.}
\label{fig:sot_enhance_OUC_haze}
\end{figure*}

Fig.~\ref{fig:sot_enhance_OUC_haze} shows qualitative comparisons on the underwater images with different haze levels spreading over limpidity, medium, and turbidity on the OUC dataset. Most of the physical model-based and model-free algorithms are effective for removing three haze levels. For example, most of the physical model-based methods including UDCP, GDCP and Blurriness can remove all the three levels of haze-effects benefits from the use of the scattering prior. Histogram also effectively removes the haze-effects, while Regression still does little about the haze-effects on the results and RedChannel generates the images with unnatural appearances. Moreover, the majority of the model-free algorithms (e.g., Fusion and Retinex) are able to deal with haze-effects while most of the deep learning based algorithms (e.g., DUIENet and CycleGAN) work little on the haze removal. This is mainly because the capability of CNN modeling the transformation between haze-like and haze-free images is limited. OurPatch and OursOF can effectively remove haze-effects as the detection perceptors help recover much details.

\section*{Supplementary I}
\begin{table*}[h]
\begin{center}
\caption{Full-Reference image quality and detection accuracy evaluations on the synthetic MultiviewUnderwater dataset.}
\label{table:quantiMultiView_perceptor}
\resizebox{\textwidth}{12mm}{
\begin{tabular}{cccccccccccccc}
\hline\noalign{\smallskip}
Methods & UDCP & GDCP & Blurriness & Regression & RedChannel & Histogram & Fusion & Twostep & Retinex & DUIENet & CycleGAN & OursPatch & OursOF\\
\noalign{\smallskip}
\hline
\noalign{\smallskip}
MSE & 1.0577 & 3.2625 & 0.9387 & 0.7245 & 0.7298 & 0.8553 & 1.1339 & 2.1455 & 0.7351 & 0.2689 & 0.7778 & \textbf{0.0453} & 0.1683\\
PSNR & 18.510 & 13.443 & 19.613 & 19.921 & 20.271 & 19.1972 & 18.232 & 15.010 & 19.858 & 23.8766 & 20.7561 & \textbf{33.3687} & 26.1421\\
SSIM & 0.1840 & 0.2710 & 0.2518 & 0.2125 & 0.4035 & 0.5479 & 0.4604 & 0.3712 & 0.5222 & 0.7768 & 0.5509 & \textbf{0.9374} & 0.6364\\
PCQI & 0.5463 & 0.5374 & 0.5965 & 0.5997 & 0.5939 & 0.5972 & 0.5775 & 0.5002 & 0.6333 & 0.6840 & 0.5558 & \textbf{0.8441} & 0.6741\\
mAP & 72.1 & 74.1 & 74.4 & 71.9 & 76.4 & 78.3 & 77.1 & 66.0 & 78.2 & 76.3 & 74.8 & 79.9 & \textbf{86.7}\\
\hline
\end{tabular}}
\end{center}
\end{table*}

\begin{table*}[h]
\begin{center}
\caption{Full-Reference image quality and detection accuracy evaluations of different UIE algorithms on the OUC dataset.}
\label{table:quantiOUC_sot}
\resizebox{\textwidth}{12mm}{
\begin{tabular}{cccccccccccccc}
\hline\noalign{\smallskip}
Methods & UDCP & GDCP & Blurriness & Regression & RedChannel & Histogram & Fusion & Twostep & Retinex & WaterGAN  & CycleGAN & OurPD & OurOFD\\
\noalign{\smallskip}
\hline
\noalign{\smallskip}
MSE & 3.3769 & 2.2945 & 0.6279 & 0.4150 & 7.2612 & 0.5640 & 0.2508 & 1.5833 & 0.4083 & 0.1217 & 0.1361 & \textbf{0.0224} & 0.0616\\
PSNR & 13.3206 & 15.9326 & 20.8400 & 22.2458 & 9.6755 & 20.9896 & 28.4754 & 16.2539 & 26.5359 & 27.9608 & 26.8538 & \textbf{35.4039} & 30.8662\\
SSIM & 0.5130 & 0.6407 & 0.7472 & 0.5691 & 0.1696 & 0.7602 & 0.8905 & 0.6002 & 0.8789 & 0.8412 & 0.8970 & \textbf{0.9724} & 0.9351\\
PCQI & 0.4259 & 0.5968 & 0.6678 & 0.6741 & 0.1524 & 0.8102 & 0.9270 & 0.4898 & 0.8337 & 0.7405 & 0.9355 & \textbf{0.9389} & 0.9030\\
mAP & 87.1 & 86.9 & 86.4 &  81.6 & 41.6 & 81.5 & 83.9 & 74.8 & 87.2 & 84.0 & 82.0 & 86.5 & \textbf{90.1}\\
\hline
\end{tabular}}
\end{center}
\end{table*}

\section*{Supplementary J}
\begin{figure*}[h]
\centering
\includegraphics[height=11.5cm, width=18cm]{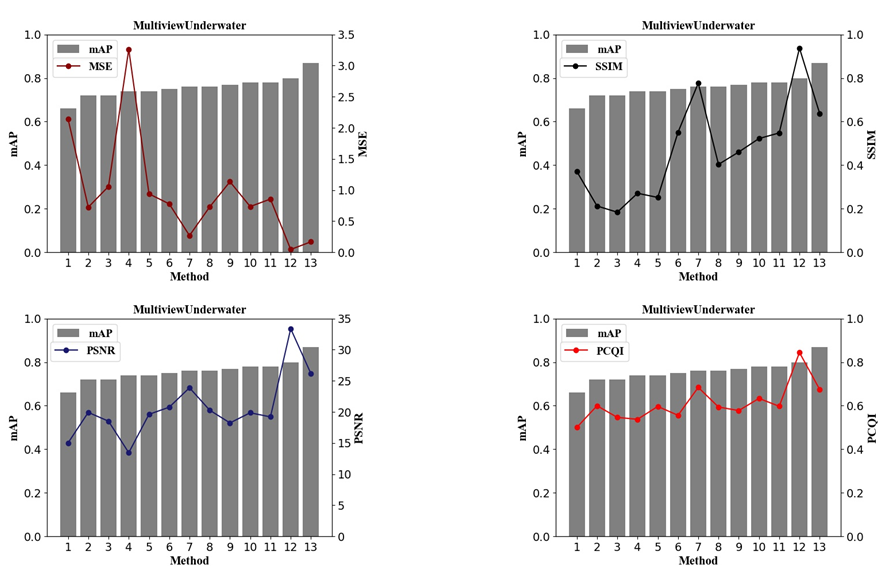}
\caption{Image quality evaluation metrics and mAP on MultiviewUnderwater. The histogram represents the mAP and the polyline represents different image quality evaluation metrics. Numbers 1 to 13 refer to thirteen UIE algorithms ordered according to increasing mAP values.}
\label{fig:metricsrelation}
\end{figure*}

\section*{Supplementary K}
\begin{figure*}[h]
\centering
\includegraphics[height=9cm, width=18cm]{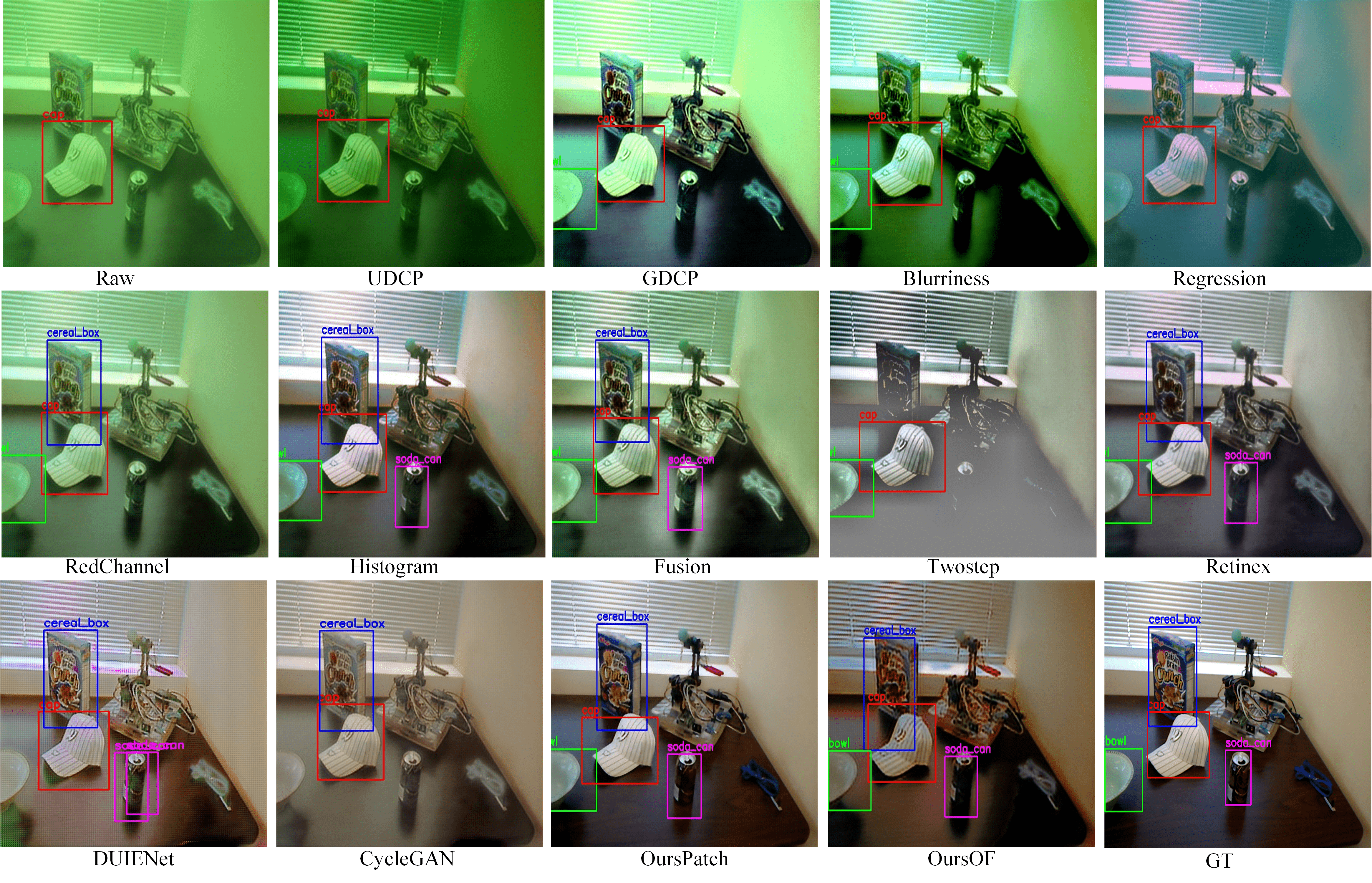}
\caption{Visualization of object detection results after having applied different UIE algorithms on MultiViewUnderwater.}
\label{fig:sot_detection_MultiView}
\end{figure*}
\begin{figure*}[h]
\centering
\includegraphics[height=9cm, width=18cm]{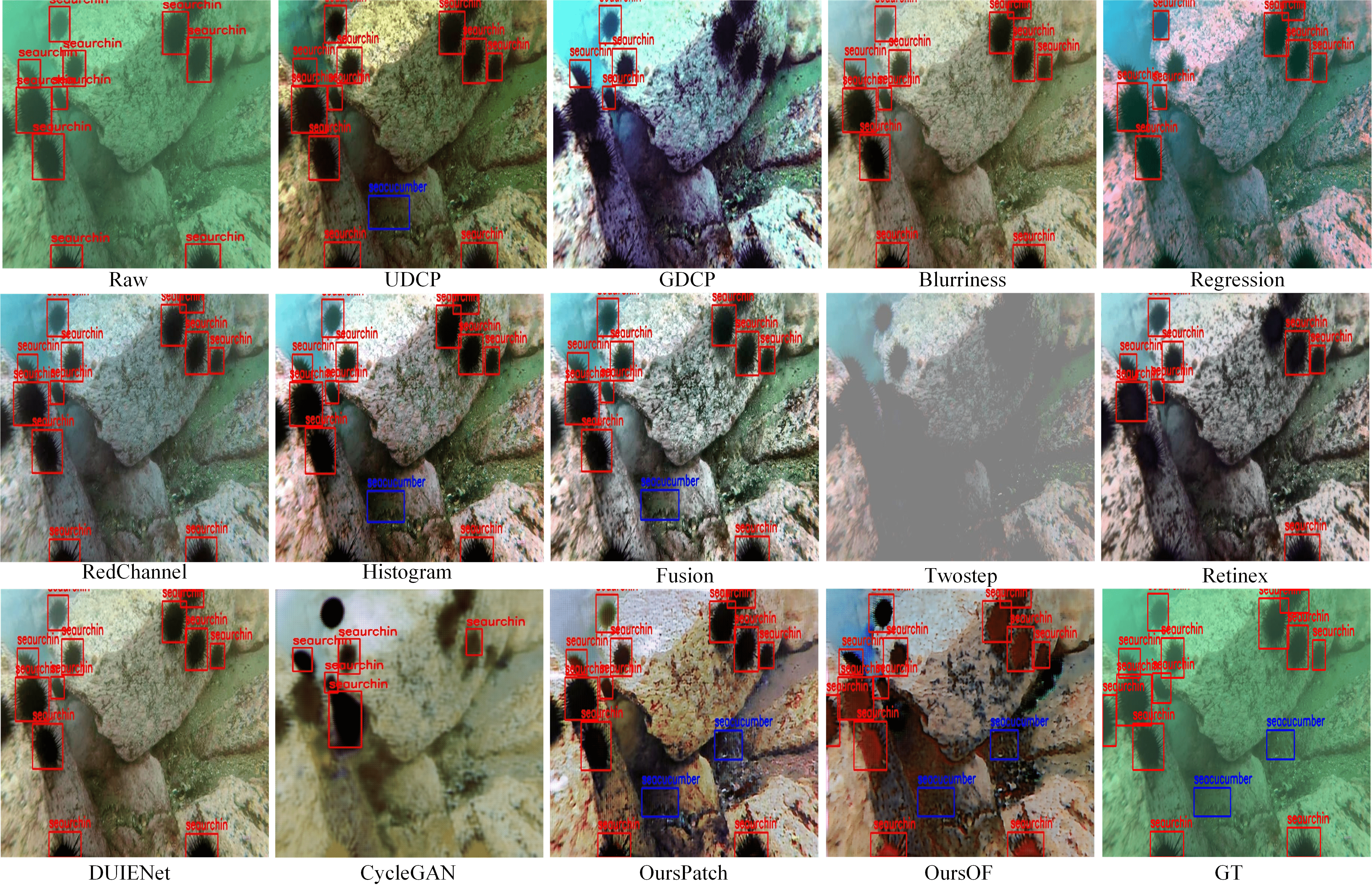}
\caption{Visualization of object detection results after having applied different UIE algorithms on ChinaMM.}
\label{fig:sot_detection_ChinaMM}
\end{figure*}


\begin{thebibliography}{00}
\bibitem{b1} Liu, R., Fan, X., Zhu, M., Hou, M., and Luo, Z. (2020). Real-world Underwater Enhancement: Challenges, Benchmarks, and Solutions under Natural Light. IEEE Transactions on Circuits and Systems for Video Technology. [Online]. Available: https://doi.org/10.1109/TCSVT.2019.2963772
\bibitem{b2} Zhou, Y., Wu, Q., Yan, K., Feng, L., and Xiang, W. (2018). Underwater image restoration using color-line model. IEEE Transactions on Circuits and Systems for Video Technology, 29(3), 907-911.
\bibitem{b3} Ye, X., Li, Z., Sun, B., Wang, Z., Xu, R., Li, H., and Fan, X. (2019). Deep Joint Depth Estimation and Color Correction from Monocular Underwater Images based on Unsupervised Adaptation Networks. IEEE Transactions on Circuits and Systems for Video Technology.
\bibitem{b4} Li, J., Skinner, K. A., Eustice, R. M., and Johnson-Roberson, M. (2017). WaterGAN: Unsupervised generative network to enable real-time color correction of monocular underwater images. IEEE Robotics and Automation letters, 3(1), 387-394.
\bibitem{b5} Fabbri, C., Islam, M. J., and Sattar, J. (2018, May). Enhancing underwater imagery using generative adversarial networks. In 2018 IEEE International Conference on Robotics and Automation (ICRA) (pp. 7159-7165). IEEE.
\bibitem{b6} Liu, W., Anguelov, D., Erhan, D., Szegedy, C., Reed, S., Fu, C. Y., and Berg, A. C. (2016, October). Ssd: Single shot multibox detector. In European Conference on Computer Vision (pp. 21-37). Springer, Cham.
\bibitem{b7} Ancuti, C., Ancuti, C. O., Haber, T., and Bekaert, P. (2012, June). Enhancing underwater images and videos by fusion. In 2012 IEEE Conference on Computer Vision and Pattern Recognition (pp. 81-88). IEEE.
\bibitem{b8} Fu, X., Zhuang, P., Huang, Y., Liao, Y., Zhang, X. P., and Ding, X. (2014, October). A retinex-based enhancing approach for single underwater image. In 2014 IEEE International Conference on Image Processing (ICIP) (pp. 4572-4576). IEEE.
\bibitem{b9} Li, C., Anwar, S., and Porikli, F. (2020). Underwater scene prior inspired deep underwater image and video enhancement. Pattern Recognition, 98, 107038.
\bibitem{b10} Peng, Y. T., and Cosman, P. C. (2017). Underwater image restoration based on image blurriness and light absorption. IEEE Transactions on Image Processing, 26(4), 1579-1594.
\bibitem{b11} Wang, Y., Zhang, J., Cao, Y., and Wang, Z. (2017, September). A deep CNN method for underwater image enhancement. In 2017 IEEE International Conference on Image Processing (ICIP) (pp. 1382-1386). IEEE.
\bibitem{b12} Chiang, J. Y., and Chen, Y. C. (2011). Underwater image enhancement by wavelength compensation and dehazing. IEEE Transactions on Image Processing, 21(4), 1756-1769.
\bibitem{b13} Berman, D., Treibitz, T., and Avidan, S. (2017, September). Diving into haze-lines: Color restoration of underwater images. In Proc. British Machine Vision Conference (BMVC) (Vol. 1, No. 2).
\bibitem{b14} Jerlov, N. G. (1976). Marine optics. Elsevier.
\bibitem{b15} Gupta, H., and Mitra, K. (2019, September). Unsupervised Single Image Underwater Depth Estimation. In 2019 IEEE International Conference on Image Processing (ICIP) (pp. 624-628). IEEE.
\bibitem{b16} Łuczynski, T., Petillot, Y., and Wang, S. Seeing through Water: From Underwater Image Synthesis to Generative Adversarial Networks based Underwater Single Image Enhancement. In: ICRA Workshop. (2019)
\bibitem{b17} Zhu, J. Y., Park, T., Isola, P., and Efros, A. A. (2017). Unpaired image-to-image translation using cycle-consistent adversarial networks. In Proceedings of the IEEE International Conference on Computer Vision (pp. 2223-2232).
\bibitem{b18} Hou, M., Liu, R., Fan, X., and Luo, Z. (2018, October). Joint residual learning for underwater image enhancement. In 2018 25th IEEE International Conference on Image Processing (ICIP) (pp. 4043-4047). IEEE.
\bibitem{b19} Zhu, J. Y., Zhang, R., Pathak, D., Darrell, T., Efros, A. A., Wang, O., and Shechtman, E. (2017). Toward multimodal image-to-image translation. In Advances in Neural Information Processing Systems (pp. 465-476).
\bibitem{b20} You, H., Cheng, Y., Cheng, T., Li, C., and Zhou, P. (2018). Bayesian CycleGAN via Marginalizing Latent Sampling. arXiv preprint arXiv:1811.07465.
\bibitem{b21} Fu, X., Fan, Z., Ling, M., Huang, Y., and Ding, X. (2017, November). Two-step approach for single underwater image enhancement. In 2017 International Symposium on Intelligent Signal Processing and Communication Systems (ISPACS) (pp. 789-794). IEEE.
\bibitem{b22} Zhang, S., Wang, T., Dong, J., and Yu, H. (2017). Underwater image enhancement via extended multi-scale Retinex. Neurocomputing, 245, 1-9.
\bibitem{b23} Drews, P. L., Nascimento, E. R., Botelho, S. S., and Campos, M. F. M. (2016). Underwater depth estimation and image restoration based on single images. IEEE Computer Graphics and Applications, 36(2), 24-35.
\bibitem{b24} Peng, Y. T., Cao, K., and Cosman, P. C. (2018). Generalization of the dark channel prior for single image restoration. IEEE Transactions on Image Processing, 27(6), 2856-2868.
\bibitem{b25} Galdran, A., Pardo, D., Picón, A., and Alvarez-Gila, A. (2015). Automatic red-channel underwater image restoration. Journal of Visual Communication and Image Representation, 26, 132-145.
\bibitem{b26} He, K., Sun, J., and Tang, X. (2010). Single image haze removal using dark channel prior. IEEE Transactions on Pattern Analysis and Machine Intelligence, 33(12), 2341-2353.
\bibitem{b27} Li, C., Guo, J., Guo, C., Cong, R., and Gong, J. (2017). A hybrid method for underwater image correction. Pattern Recognition Letters, 94, 62-67.
\bibitem{b28} Li, C. Y., Guo, J. C., Cong, R. M., Pang, Y. W., and Wang, B. (2016). Underwater image enhancement by dehazing with minimum information loss and histogram distribution prior. IEEE Transactions on Image Processing, 25(12), 5664-5677.
\bibitem{b29} C. Li, C. Guo, W. Ren, R. Cong, J. Hou, S. Kwong, and D. Tao. An Underwater Image Enhancement Benchmark Dataset and Beyond. IEEE Transactions on Image Processing., vol. 29, pp.4376-4389, 2019.
\bibitem{b30} Ye, X., Xu, H., Ji, X., and Xu, R. (2018, September). Underwater Image Enhancement Using Stacked Generative Adversarial Networks. In Pacific Rim Conference on Multimedia (pp. 514-524). Springer, Cham.
\bibitem{b31} Szegedy, C., Zaremba, W., Sutskever, I.. Bruna, J., Erhan, D., Goodfellow, I., and Fergus, R.: Intriguing properties of neural networks. In: ICLR. (2014)
\bibitem{b32} Nguyen, A., Yosinski, J., and Clune, J.. Deep neural networks are easily fooled: High confidence predictions for unrecognizable images. In: CVPR. (2015)
\bibitem{b33} Johnson, J., Alahi, A., and Fei-Fei, L. (2016, October). Perceptual losses for real-time style transfer and super-resolution. In European Conference on Computer Vision (pp. 694-711). Springer, Cham.
\bibitem{b34} Dosovitskiy, A., and Brox, T. (2016). Generating images with perceptual similarity metrics based on deep networks. In Advances in Neural Information Processing Systems (pp. 658-666).
\bibitem{b35} Estrach, J. B., Sprechmann, P., and LeCun, Y. (2016, January). Super-resolution with deep convolutional sufficient statistics. In 4th International Conference on Learning Representations, ICLR 2016.
\bibitem{b36} Zhang, R., Isola, P., Efros, A. A., Shechtman, E., and Wang, O. (2018). The unreasonable effectiveness of deep features as a perceptual metric. In Proceedings of the IEEE Conference on Computer Vision and Pattern Recognition (pp. 586-595).
\bibitem{b37} Simonyan, K., Vedaldi, A., and Zisserman, A.. Deep inside convolutional networks: Visualising image classification models and saliency maps. In: ICLR Workshop. (2014)
\bibitem{b38} Yosinski, J., Clune, J., Nguyen, A., Fuchs, T., and Lipson, H.. Understanding neural networks through deep visualization. In: ICML Deep Learning Workshop. (2015)
\bibitem{b39} Lai, K., Bo, L., Ren, X., and Fox, D. (2011, May). A large-scale hierarchical multi-view rgb-d object dataset. In 2011 IEEE International Conference on Robotics and Automation (pp. 1817-1824). IEEE.
\bibitem{b40} Jian, M., Qi, Q., Dong, J., Yin, Y., Zhang, W., and Lam, K. M. (2017, July). The OUC-vision large-scale underwater image database. In 2017 IEEE International Conference on Multimedia and Expo (ICME) (pp. 1297-1302). IEEE.
\bibitem{b41} Wang, S., Ma, K., Yeganeh, H., Wang, Z., and Lin, W. (2015). A patch-structure representation method for quality assessment of contrast changed images. IEEE Signal Processing Letters, 22(12), 2387-2390.
\bibitem{b42} Panetta, K., Gao, C., and Agaian, S. (2015). Human-visual-system-inspired underwater image quality measures. IEEE Journal of Oceanic Engineering, 41(3), 541-551.
\bibitem{b43} Yang, M., and Sowmya, A. (2015). An underwater color image quality evaluation metric. IEEE Transactions on Image Processing, 24(12), 6062-6071.
\bibitem{b44} Kingma, D. P., and Ba, J. L. (2015). Adam: A method for stochastic gradient descent. In ICLR: International Conference on Learning Representations.
\bibitem{b45} Hoiem, D., Chodpathumwan, Y., and Dai, Q. (2012, October). Diagnosing error in object detectors. In European Conference on Computer Vision (pp. 340-353). Springer, Berlin, Heidelberg.
\end{thebibliography}
\end{document}